\documentclass[11pt]{article}

\usepackage{acl}
\usepackage{times}
\usepackage{latexsym}

\usepackage[T1]{fontenc}

\usepackage[utf8]{inputenc}

\usepackage{microtype}

\usepackage{inconsolata}

\usepackage{graphicx}

\usepackage{booktabs}
\usepackage{multirow}
\usepackage{amsmath}
\usepackage{amssymb}
\usepackage{url}
\usepackage{enumitem}

\usepackage{algorithm}
\usepackage{algpseudocode}

\usepackage{siunitx}

\usepackage{float}

\usepackage{listings}

\usepackage{tcolorbox}
\tcbuselibrary{breakable,skins,listings}

\usepackage{caption}
\tcbset{
  promptbox/.style={
    enhanced,
    colback=gray!3,
    breakable,
    colframe=gray!60,
    boxrule=0.5pt,
    arc=2pt,
    left=6pt, right=6pt, top=6pt, bottom=6pt,
    fontupper=\small\ttfamily,
    before upper={\parindent0pt\parskip4pt}
  },
  systemprompt/.style={
    enhanced,
    colback=gray!3,
    breakable,
    colframe=gray!70,
    coltitle=black,
    fonttitle=\small\bfseries,
    boxrule=0.5pt,
    arc=2pt,
    left=6pt, right=6pt, top=6pt, bottom=6pt,
    fontupper=\small\ttfamily,
    before upper={\parindent0pt\parskip4pt},
    attach boxed title to top left={yshift=-2mm, xshift=4mm},
    boxed title style={colback=gray!20, colframe=gray!70, arc=1pt, boxrule=0.3pt}
  },
  humanprompt/.style={
    enhanced,
    colback=blue!2,
    breakable,
    colframe=blue!40,
    coltitle=black,
    fonttitle=\small\bfseries,
    boxrule=0.5pt,
    arc=2pt,
    left=6pt, right=6pt, top=6pt, bottom=6pt,
    fontupper=\small\ttfamily,
    before upper={\parindent0pt\parskip4pt},
    attach boxed title to top left={yshift=-2mm, xshift=4mm},
    boxed title style={colback=blue!10, colframe=blue!40, arc=1pt, boxrule=0.3pt}
  },
  assistantprompt/.style={
    enhanced,
    colback=green!2,
    breakable,
    colframe=green!40!black,
    coltitle=black,
    fonttitle=\small\bfseries,
    boxrule=0.5pt,
    arc=2pt,
    left=6pt, right=6pt, top=6pt, bottom=6pt,
    fontupper=\small\ttfamily,
    before upper={\parindent0pt\parskip4pt},
    attach boxed title to top left={yshift=-2mm, xshift=4mm},
    boxed title style={colback=green!10, colframe=green!40!black, arc=1pt, boxrule=0.3pt}
  },
  rubricbox/.style={
    enhanced,
    colback=yellow!3,
    breakable,
    colframe=yellow!50!black,
    coltitle=black,
    fonttitle=\small\bfseries,
    boxrule=0.5pt,
    arc=2pt,
    left=6pt, right=6pt, top=6pt, bottom=6pt,
    fontupper=\small\ttfamily,
    before upper={\parindent0pt\parskip4pt},
    attach boxed title to top left={yshift=-2mm, xshift=4mm},
    boxed title style={colback=yellow!15, colframe=yellow!50!black, arc=1pt, boxrule=0.3pt}
  },
  codebox/.style={
    enhanced,
    colback=gray!3,
    breakable,
    colframe=gray!50,
    coltitle=black,
    fonttitle=\small\bfseries,
    boxrule=0.5pt,
    arc=2pt,
    left=4pt, right=4pt, top=4pt, bottom=4pt,
    fontupper=\scriptsize\ttfamily,
    before upper={\parindent0pt\parskip2pt},
    attach boxed title to top left={yshift=-2mm, xshift=4mm},
    boxed title style={colback=gray!15, colframe=gray!50, arc=1pt, boxrule=0.3pt}
  }
}

\lstdefinestyle{pythonstyle}{
  language=Python,
  basicstyle=\scriptsize\ttfamily,
  keywordstyle=\color{blue!70!black},
  commentstyle=\color{green!50!black},
  stringstyle=\color{orange!70!black},
  showstringspaces=false,
  breaklines=true,
  frame=none,
  numbers=none,
  aboveskip=2pt,
  belowskip=2pt
}

\setlist[itemize]{leftmargin=*, nosep}
\setlist[enumerate]{leftmargin=*, nosep}

\title{From XAI to Stories: A Factorial Study of LLM-Generated Explanation Quality}

\author{
    \textbf{Fabian Lukassen}$^{1}$,
    \textbf{Jan Herrmann}$^{2}$,
    \textbf{Christoph Weisser}$^{4}$,
    \textbf{Benjamin Saefken}$^{1,3}$,
    \textbf{Thomas Kneib}$^{1}$\\
    $^{1}$University of G\"ottingen \quad $^{2}$BASF SE \quad $^{3}$TU Clausthal \quad $^{4}$ Hochschule Bielefeld \\
    \texttt{fabian.lukassen@stud.uni-goettingen.de}, \texttt{jan.herrmann@basf.com}, \texttt{christoph.weisser@hsbi.de} \\
    \texttt{benjamin.saefken@tu-clausthal.de}, \texttt{tkneib@uni-goettingen.de}
}

\begin{document}
\maketitle

\begin{abstract}
Explainable AI (XAI) methods like SHAP and LIME produce numerical feature attributions that remain inaccessible to non-expert users. Prior work has shown that Large Language Models (LLMs) can transform these outputs into natural language explanations (NLEs), but it remains unclear which factors contribute to high-quality explanations. We present a systematic factorial study investigating how Forecasting model choice, XAI method, LLM selection, and prompting strategy affect NLE quality. Our design spans four models (XGBoost (XGB), Random Forest (RF), Multilayer Perceptron (MLP), and SARIMAX -- comparing black-box Machine-Learning (ML) against classical time-series approaches), three XAI conditions (SHAP, LIME, and a \textit{no-XAI baseline}), three LLMs (GPT-4o, Llama-3-8B, DeepSeek-R1), and eight prompting strategies. Using G-Eval, an LLM-as-a-judge evaluation method, with dual LLM judges and four evaluation criteria, we evaluate 660 explanations for time-series forecasting. Our results suggest that: (1) XAI provides only small improvements over no-XAI baselines, and only for expert audiences; (2) LLM choice dominates all other factors, with DeepSeek-R1 outperforming GPT-4o and Llama-3; (3) we observe an \textit{interpretability paradox}: in our setting, SARIMAX yielded lower NLE quality than ML models despite having higher prediction accuracy and being inherently interpretable; (4) zero-shot prompting is competitive with self-consistency at 7$\times$ lower cost; and (5) chain-of-thought hurts rather than helps.\footnote{Code and data: \url{https://anonymous.4open.science/r/DOE_XAI_LLM-9FBD/}}
\end{abstract}

\begin{figure}[H]
\centering
\includegraphics[width=0.48\textwidth]{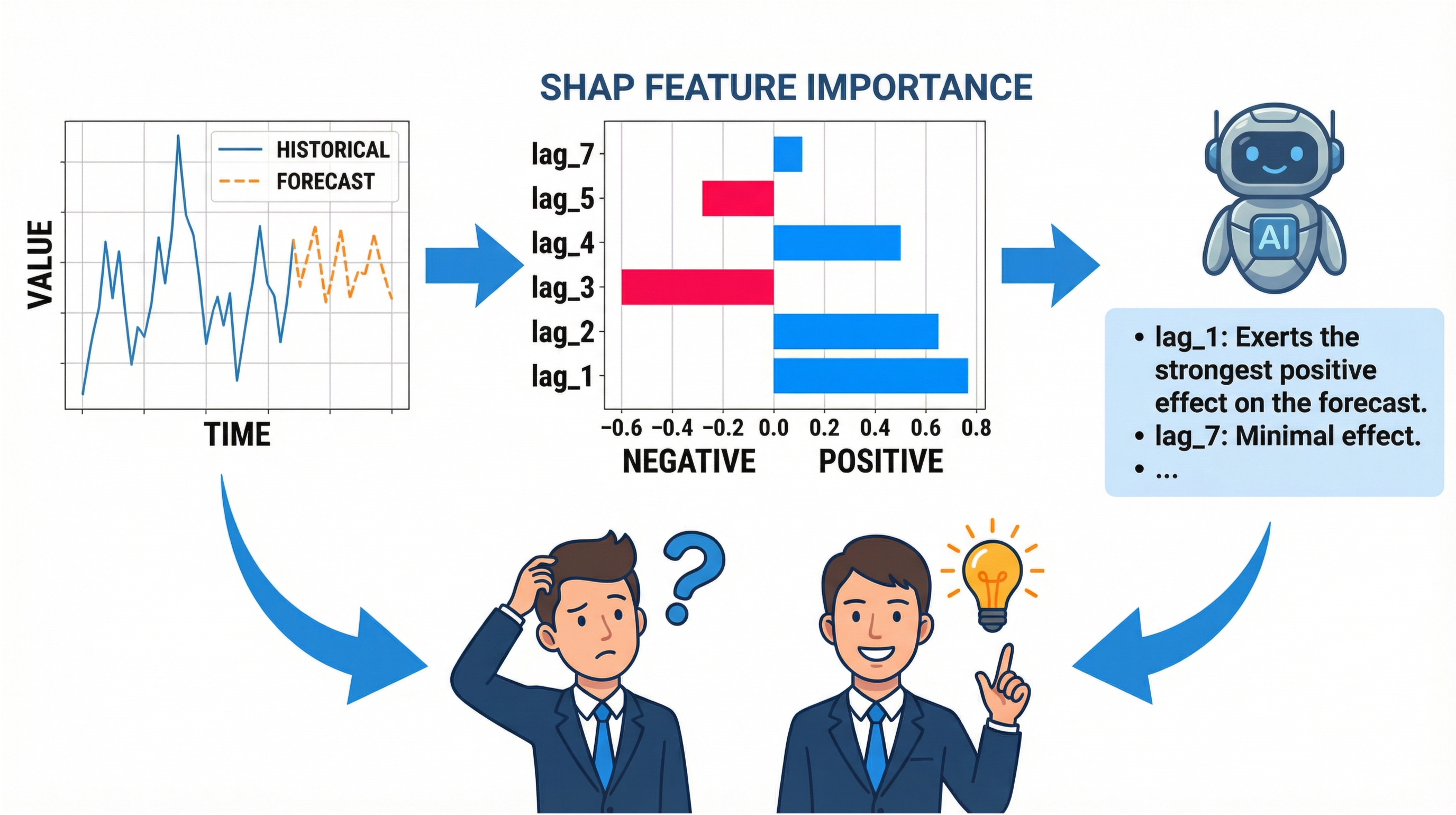}
\caption{Overview of our Prediction-to-NLE pipeline: Models generate predictions, XAI methods (SHAP/LIME) compute feature attributions, and LLMs transform these numerical outputs into narratives accessible to lay users.}
\label{fig:workflow}
\end{figure}

\section{Introduction}

The deployment of ML models in critical domains such as healthcare, finance, and energy management has intensified the need for model interpretability \citep{guidotti2018survey}. XAI methods like SHAP \citep{lundberg2017unified} and LIME \citep{ribeiro2016why} have emerged as essential tools, providing quantitative feature attributions that reveal which inputs drive model predictions. However, these technical outputs -- typically numerical importance scores or attribution values -- remain inaccessible to non-expert users who lack the statistical background to interpret them \citep{zytek2024llms}.

Consider a model predicting next week's household energy consumption. A typical SHAP output might show: ``\texttt{lag\_1: +0.23, weekofyear: -0.15, holiday\_count: +0.08}.'' While technically precise, a homeowner cannot readily understand that last week's high usage is pushing the forecast up, that the current season typically has lower demand, or that an upcoming holiday slightly increases expected consumption. This interpretability gap limits the practical impact of XAI methods and hinders the democratization of model transparency.

Recent work has demonstrated that LLMs can transform XAI output into NLEs that non-experts can better understand (Figure~\ref{fig:workflow}), with user studies validating this approach \citep{zytek2024explingo, cedro2024graphxain}.

\begin{figure}[h]
    \centering
    \includegraphics[width=0.5\linewidth]{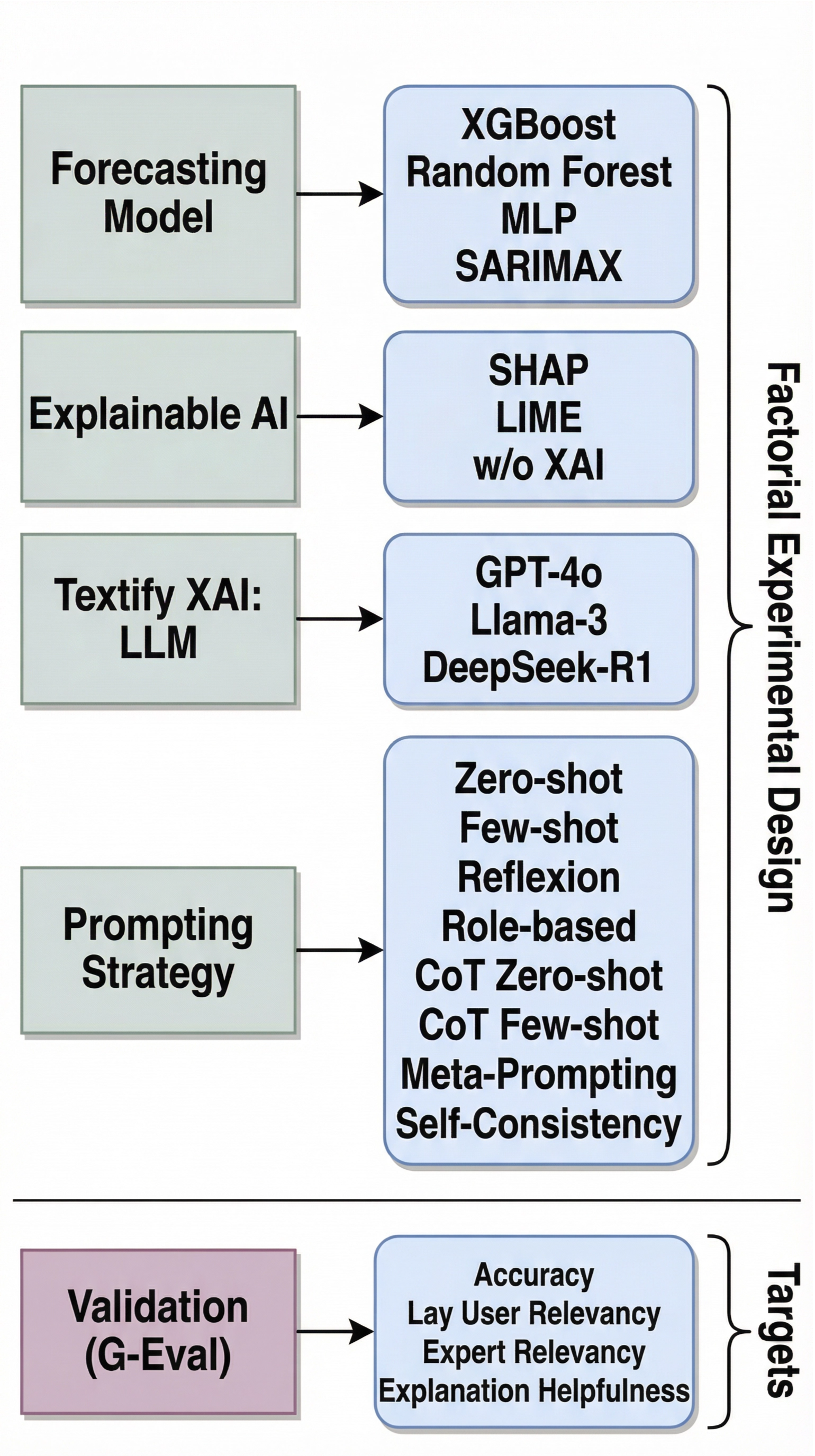}
    \caption{Factorial experimental design with four factors and their levels (top), and five G-Eval evaluation criteria (bottom).}
    \label{fig:factorial_design}
\end{figure}

However, previous work on transforming XAI outputs into NLEs does not systematically study the effects of ML model choice, XAI method, LLM, and prompting strategy on NLE quality. We address this gap with a factorial study grounded in Design of Experiments (DOE) principles \citep{collins2014factorial}, focusing on time-series forecasting. A domain where temporal dependencies and seasonal patterns make XAI outputs especially challenging to interpret.
The factorial combination of these four factors (see Figure~\ref{fig:factorial_design} for their levels) yields 660 NLEs, which we evaluate using G-Eval \citep{liu2023geval}, an LLM-as-judge framework with dual judges and four evaluation criteria. Our main contributions are:

\begin{enumerate}
    \item A \textbf{factorial study design} with statistical analysis (ANOVA, effect sizes, robustness checks) to systematically quantify what factors matter for XAI-to-NLE quality, including a crucial \textit{no-XAI baseline} absent from prior work.
    \item The observation of an \textbf{``interpretability paradox''}: in our setting, SARIMAX yielded lower NLE quality than ML models despite higher prediction accuracy and inherent interpretability.
    \item \textbf{Empirical evidence that simplicity suffices}: zero-shot prompting achieves near-equivalent quality to complex strategies at 7$\times$ lower cost, and XAI provides only modest improvements ($\omega^2=.02$) over no-XAI baselines.
    \item \textbf{Evidence that LLM choice dominates}: LLM selection explains far more variance in NLE quality than XAI method, prompting strategy, or ML model choice.
\end{enumerate}

\section{Related Work}

\subsection{Explainable AI Methods}

Among post-hoc explanation methods, SHAP and LIME have emerged as the two most widely adopted techniques for interpreting black-box models \citep{garreau2020explaining}.

SHAP \citep{lundberg2017unified} provides theoretically grounded attributions based on Shapley values from cooperative game theory. For a prediction $f(x)$ with feature set $\mathcal{F}$, SHAP assigns each feature $i$ an importance value $\phi_i$ computed as a weighted average of marginal contributions across all possible feature coalitions $S$:
\begin{equation}
\phi_i = \sum_{S \subseteq \mathcal{F} \setminus \{i\}} w_{|S|}  \left[ f(x_{S \cup \{i\}}) - f(x_S) \right]
\end{equation}
where $w_{|S|} = \frac{|S|!(|\mathcal{F}|-|S|-1)!}{|\mathcal{F}|!}$. The explanation satisfies local accuracy: $f(x) = \phi_0 + \sum_{i=1}^{M} \phi_i$, with $\phi_0$ being the baseline prediction.

LIME \citep{ribeiro2016why} takes a different approach, fitting interpretable local surrogate models. Given instance $x$ and black-box model $f$, LIME learns a locally faithful linear model $g$ by solving:
\begin{equation}
\xi(x) = \arg\min_{g \in \mathcal{G}} \mathcal{L}(f, g, \pi_x) + \Omega(g)
\end{equation}
where $\pi_x$ is a proximity measure defining the local neighborhood and $\Omega(g)$ penalizes complexity.

Both methods have been widely adopted \citep{guidotti2018survey}, applied in complex machine learning pipelines \citep{semnani2024amachine, semnani2025from}, and extended to specialized variants including TreeSHAP \citep{lundberg2020local}. However, despite their mathematical rigor, XAI outputs require statistical expertise to interpret, particularly for time-series data where temporal dependencies add complexity \citep{theissler2022explainable}. \citet{zytek2024llms} identify this as a fundamental barrier, noting that ``feature importance scores are only useful if users can understand what they mean.''

In contrast, classical statistical models like SARIMAX offer inherent interpretability through coefficients and confidence intervals \citep{box2015time}.

\subsection{Natural Language Explanations from XAI}

The transformation of XAI outputs to natural language has progressed from template-based systems to LLM-based approaches. \citet{martens2024xaistories} introduced XAIstories, generating narratives from SHAP values, with user studies showing participants found them convincing and useful for communicating with laypeople. \citet{zytek2024explingo} developed Explingo, also using SHAP, combining XAI-to-narrative transformation with an LLM-based grader. \citet{cedro2024graphxain} extended this to graph neural networks using GNNExplainer. For time-series specifically, \citet{shajalal2024forecastexplainer} used DeepLIFT for LSTM-based energy forecasting, while \citet{aksu2024xforecast} proposed XForecast with simulatability metrics, finding that an LLM's numerical reasoning ability matters more than model size for accurate explanations. Domain applications include chemistry \citep{wellawatte2023extracting}.

\subsection{LLM Prompting Strategies}

The effectiveness of LLMs depends heavily on prompting. Foundational techniques include \textbf{zero-shot} and \textbf{few-shot} in-context learning \citep{brown2020language}, while more advanced strategies target reasoning and reliability. \textbf{Chain-of-thought (CoT)} prompting elicits step-by-step reasoning \citep{wei2022chain}, which \textbf{self-consistency} extends by sampling multiple reasoning paths and selecting the most common answer \citep{wang2023selfconsistency}. Other approaches include \textbf{reflexion}, where models iteratively critique and refine their outputs \citep{shinn2023reflexion}; \textbf{meta-prompting}, which uses structured templates for multi-step reasoning \citep{suzgun2024metaprompting}; and \textbf{role-based} prompting, which assigns personas to guide generation \citep{kong2024betterzeroshotreasoning}.

\subsection{LLM-as-Judge Evaluation}
\label{sec:llm-judge}

Traditional metrics (BLEU, ROUGE) correlate poorly with human judgment for open-ended text generation \citep{liu2023geval}. The LLM-as-judge paradigm has emerged as a scalable alternative \citep{zheng2023judging}, with three main evaluation paradigms \citep{gu2024llmasjudge}: \textit{pointwise} (scoring individual outputs independently), \textit{pairwise} (comparing two outputs to select the better one), and \textit{listwise} (ranking multiple outputs simultaneously).

\citet{liu2023geval} introduced G-Eval, a pointwise framework combining chain-of-thought prompting with probability-weighted scoring. G-Eval first generates evaluation steps from a criterion (e.g., ``coherence''), then uses these steps to assess outputs via a form-filling paradigm. To obtain fine-grained scores, G-Eval computes:
{\setlength{\abovedisplayskip}{4pt}
\setlength{\belowdisplayskip}{4pt}
\begin{equation}
\text{Score}_{\text{G-Eval}}(y) = \sum_{i=1}^{N} i \cdot P(\text{score}=i | y, c, d)
\end{equation}}%
where $P(\text{score}=i)$ is derived from the softmax over token log-probabilities for score tokens ``1'' through ``$N$'', $y$ is the generated text, $c$ is the evaluation context, and $d$ is the dimension-specific rubric. This yields continuous scores (e.g., 3.7) rather than discrete integers, achieving 0.514 Spearman correlation with human judgments.

However, LLM judges exhibit systematic biases \citep{gu2024llmasjudge}: \textit{positional bias} (preference for first or last positions in ranking tasks), \textit{self-preference bias} (favoring outputs from the same LLM family), and \textit{verbosity bias} (favoring longer responses). Mitigation strategies include randomizing presentation order, using judges from different model families than generators, and supplementing absolute scores with ranking-based evaluation.

\subsection{Positioning Our Work}

While previous work demonstrates XAI-to-NLE feasibility, none systematically investigates which factors matter. We fill this gap with a factorial design that varies ML models (including classical vs.\ ML comparison), XAI methods (including no-XAI baseline), LLMs, and prompting strategies. This enables us to address the following research questions (RQs):

\begin{itemize}
    \item \textbf{RQ1 (XAI Value):} Does including XAI outputs improve NLE quality compared to no-XAI baselines, and do SHAP and LIME differ in their effectiveness?
    \item \textbf{RQ2 (LLM Characteristics):} Does the choice of the LLM (varying in scale and reasoning capabilities) affect explanation quality?
    \item \textbf{RQ3 (Model Effects):} Do different ML models and their predictive quality affect NLE quality?
    \item \textbf{RQ4 (Model Type):} Do classical time-series models yield different explanation quality than black-box ML models?
    \item \textbf{RQ5 (Prompting):} Which prompting strategy is most effective?
\end{itemize}

\section{Methodology}

\subsection{Problem Formulation}

Algorithm~\ref{alg:pipeline} outlines our Prediction-to-evaluated-NLE pipeline. 

\vspace{-2mm}
\begin{algorithm}[H]
\small
\caption{Experimental Pipeline}
\label{alg:pipeline}
\begin{algorithmic}[1]
\Require Dataset $\mathcal{D}$, Models $\mathcal{M}$, XAI $\mathcal{X}$, LLMs $\mathcal{L}$, Strategies $\mathcal{S}$
\Ensure Evaluated explanation corpus $\mathcal{E}$
\State Preprocess $\mathcal{D}$; split train/test chronologically
\Statex \textit{// Phase 1: Generation}
\ForAll{$m \in \mathcal{M}$}
    \State Train $m$; compute $\hat{\mathbf{y}}$ and metrics $(R^2, \text{MAE}, \text{RMSE})$
    \ForAll{$\chi \in \mathcal{X}$}
        \State Compute attributions $\boldsymbol{\phi} \gets \chi(m, \mathbf{z})$
        \ForAll{$\ell \in \mathcal{L}$, $s \in \mathcal{S}$}
            \State $E \gets \text{GenerateNLE}(\ell, s, \hat{y}, \boldsymbol{\phi}, \mathbf{c})$
            \State Store $(m, \chi, \ell, s, E)$ in corpus $\mathcal{C}$
        \EndFor
    \EndFor
\EndFor
\Statex \textit{// Phase 2: Evaluation}
\ForAll{$(m, \chi, \ell, s, E) \in \mathcal{C}$}
    \State $\mathbf{q} \gets \text{Evaluate}(E)$ \Comment{G-Eval scoring}
    \State Store $(m, \chi, \ell, s, E, \mathbf{q})$ in $\mathcal{E}$
\EndFor
\State \Return $\mathcal{E}$
\end{algorithmic}
\end{algorithm}
\vspace{-2mm}

Let $\mathbf{X} = \{x_t\}_{t=1}^n$ be a univariate time series. Given a forecasting model $f$ trained on features $\mathbf{z}_t$ (lagged values and covariates), we produce one-step-ahead predictions $\hat{y}_{t+1} = f(\mathbf{z}_t)$. An XAI method $\chi \in \{\text{SHAP}, \text{LIME}, \varnothing\}$ computes feature attributions $\boldsymbol{\phi}_{t+1} = \chi(f, \mathbf{z}_t)$, where $\varnothing$ denotes no XAI (attribution-free baseline).

Tuples $\mathcal{R} = \{(\hat{y}_i, \boldsymbol{\phi}_i, E_i, \mathbf{q}_i)\}_{i=1}^N$ are generated in the pipeline, where $E_i = \text{LLM}(\mathcal{P}; \hat{y}_i, \boldsymbol{\phi}_i, \mathbf{c})$ is the natural language explanation produced by prompting an LLM with strategy $\mathcal{P}$ and context $\mathbf{c}$ (model metrics, feature values), and $\mathbf{q}_i$ is the quality score from LLM-based evaluation. See Appendix~\ref{app:examples} for a complete example of input context and generated explanation.

Our factorial design spans four factors (see Figure~\ref{fig:factorial_design}). Below we detail each factor and our evaluation framework; statistical analysis follows in Section~\ref{sec:results}.

\subsection{Time-Series Forecasting Task}

We use the UCI Individual Household Electric Power Consumption dataset \citep{hebrail2012individual}, containing 2,075,259 minute-level measurements of a French household's electricity usage from December 2006 to November 2010. We chose this domain as a representative time-series forecasting task with connection to everyday decision-making: patterns are intuitive (e.g., higher consumption in winter due to heating) and LLMs can draw on common knowledge to produce meaningful explanations.

\paragraph{Data and Features.} We aggregate minute-level readings to weekly totals ($\approx$200 data points), balancing noise reduction with preserving temporal patterns. Features include lag values (past 7 weeks), ISO week number for seasonality, and holiday counts. We use a chronological 70/30 train/test split. See Appendix~\ref{app:experimental} for preprocessing details.

\subsection{Forecasting Models}

\vspace{-4mm}
\begin{table}[H]
\centering
\small
\sisetup{table-format=1.2}
\begin{tabular}{l S[table-format=1.2] S[table-format=2.2] S[table-format=2.2]}
\toprule
\textbf{Model} & {$\mathbf{R^2}$} & {\textbf{MAE}} & {\textbf{RMSE}} \\
\midrule
XGBoost & 0.69 & 20.55 & 25.03 \\
SARIMAX & 0.55 & 20.12 & 29.91 \\
Random Forest & 0.37 & 28.74 & 35.44 \\
MLP & 0.21 & 33.29 & 39.69 \\
\bottomrule
\end{tabular}
\caption{Test-set performance metrics (sorted by R$^2$).}
\label{tab:model_perf}
\end{table}

We train three ML models (XGB, RF, MLP) that deliberately vary in prediction quality (Table~\ref{tab:model_perf}), allowing us to test whether model performance affects NLE quality. We also include SARIMAX as a classical time-series counterpart with inherent interpretability through coefficients (no post-hoc XAI needed).

All models produce \textbf{one-step-ahead forecasts} under identical conditions: each prediction uses true observed values from all preceding weeks, not recursive model outputs. For ML models, lag features are pre-computed from actual consumption; for SARIMAX, we use a rolling approach that appends true values after each forecast. This ensures fair comparison across model types. See Appendix~\ref{app:experimental} for hyperparameters and visualizations.

\paragraph{Test Instance Selection.}
To ensure representative coverage across the test period, we select three specific weeks using a tercile-based sampling strategy. We divide the 52-week test set into three equal segments (early, middle, late) and select one week from each tercile where absolute prediction errors rank consistently: $|\text{XGB}| < |\text{SARIMAX}| < |\text{RF}| < |\text{MLP}|$. This criterion ensures temporal diversity while focusing on instances where model quality differences are clearly expressed, with prediction errors ranging from $\approx$4\% (XGBoost) to $\approx$36\% (MLP). See Appendix~\ref{app:experimental} for specific instance details. For each of the three weeks and each model, we generate explanations across all XAI conditions, LLMs, and prompting strategies.

\subsection{XAI Methods}

For each ML model, we compute explanations using three conditions: \textbf{SHAP} (TreeExplainer for tree-based models, KernelExplainer for MLP), \textbf{LIME} and \textbf{None} (only prediction context without feature attributions). For SARIMAX, we use only the ``none'' condition since it provides inherent interpretability through coefficients. See Appendix~\ref{app:experimental} for implementation details.

\subsection{Prompting Strategies}

We select eight prominent strategies from the literature \citep{sahoo2024systematic}, spanning simple to complex approaches: zero-shot and few-shot (direct instruction with and without examples), chain-of-thought (zero- and few-shot variants, excluding DeepSeek-R1 which natively uses CoT), meta-prompting (structured 6-step template), reflexion (iterative self-critique), role-based (expert persona), and self-consistency (ensemble of multiple outputs). These range from single-pass methods to multi-iteration approaches. All prompts share a common structure: a system message establishing the task (help a non-technical user understand the forecast), and a human message providing context including model performance metrics, the predicted value, instance features, and XAI attributions when applicable. See Appendix~\ref{app:prompts} for complete templates.

We constrain explanations to $\leq$200 words and $\leq$6 bullet points, reflecting realistic user attention spans based on cognitive load research \citep{mark2023attention, brysbaert2019reading} and controlling for verbosity bias (Section~\ref{sec:llm-judge}).

\subsection{LLMs for NLE Generation}

We use three LLMs spanning different scales and architectures: \textbf{Llama-3-8B} \citep{grattafiori2024llama}, a small open-source model run locally; \textbf{GPT-4o} \citep{openai2024gpt4o}, a large commercial model; and \textbf{DeepSeek-R1} \citep{guo2025deepseekr1}, a 671B-parameter Mixture-of-Experts model with 37B active parameters, optimized for reasoning via reinforcement learning. This selection examines how model scale and reasoning specialization affect explanation quality. We use temperature $\tau = 1.0$ for all strategies to encourage diverse outputs, except self-consistency's synthesis step ($\tau = 0.2$) which requires deterministic aggregation of multiple drafts into a single cohesive explanation \citep{wang2023selfconsistency}.

\subsection{Evaluation Framework}
\label{sec:rubrics}

We employ G-Eval \citep{liu2023geval} for pointwise scoring with two judge models: GPT-4 and DeepSeek-R1. We exclude Llama as a judge due to its smaller parameter count -- research shows that judge capability scales with model size, and weaker models produce less reliable evaluations \citep{gu2024llmasjudge}. Each dimension uses chain-of-thought evaluation steps followed by probability-weighted scoring (see Section~\ref{sec:llm-judge}). We define four evaluation dimensions grounded in XAI evaluation literature \citep{hoffman2018metrics, miller2019explanation}:

\begin{itemize}
    \item \textbf{Accuracy}: Faithfulness to provided XAI values -- correct feature signs and approximate magnitudes.
    \item \textbf{Lay User Relevancy}: Understandability for non-technical users. \citet{miller2019explanation} notes that ``practitioner satisfaction is fundamental to effective explanatory strategies.''
    \item \textbf{Expert Relevancy}: Technical soundness for ML experts. \citet{ribeiro2016why} assert that ``good explanations should provide additional insights'' enabling model assessment.
    \item \textbf{Explanation Helpfulness}: Overall pragmatic usefulness. \citet{miller2019explanation} notes individuals evaluate explanations ``according to pragmatic usefulness.''
\end{itemize}

Each rubric uses a 1--5 scale with detailed criteria (see Appendix~\ref{app:rubrics}). Following best practices from \citet{gu2024llmasjudge}, scores are averaged across both judges to mitigate self-preference bias.

\subsection{Corpus Summary}

The factorial design yields the following corpus:
\begin{itemize}
    \item \textbf{Main study:} 3 test instances $\times$ 4 models $\times$ 3 XAI $\times$ 3 LLMs $\times$ 8 strategies
    \item \textbf{Exclusions:} CoT strategies for DeepSeek-R1 (native reasoning); SHAP/LIME for SARIMAX (classical model)
    \item \textbf{Main corpus:} 660 explanations
\end{itemize}

\noindent Table~\ref{tab:efficiency} summarizes generation efficiency (tokens and time) across all experimental factors.

\begin{table}[t]
  \centering
  \small
  \begin{tabular}{@{}llcrr@{}}
    \toprule
    \textbf{Factor} & \textbf{Level} & \textbf{N} & \textbf{Tokens} & \textbf{Time (s)} \\
     & & & \tiny(653$\pm$1188) & \tiny(35.2$\pm$72.0) \\
    \midrule

    \multirow{8}{*}{\rotatebox{90}{\textbf{Strategy}}}
      & CoT Few           &   60 & \textbf{215}\tiny$\pm$40     & 35.2\tiny$\pm$34.0 \\
      & CoT Zero          &   60 & 280\tiny$\pm$46              & 29.1\tiny$\pm$27.6 \\
      & Zero-shot         &   90 & 499\tiny$\pm$618             & \textbf{22.9}\tiny$\pm$23.7 \\
      & Few-shot          &   90 & 520\tiny$\pm$743             & 29.5\tiny$\pm$31.8 \\
      & Meta-prompt       &   90 & 525\tiny$\pm$435             & 29.4\tiny$\pm$32.3 \\
      & Role-based        &   90 & 539\tiny$\pm$606             & 25.1\tiny$\pm$26.3 \\
      & Reflexion         &   90 & 2184\tiny$\pm$2869           & 132.9\tiny$\pm$190 \\
      & Self-consistency  &   90 & 3718\tiny$\pm$3446           & 219.2\tiny$\pm$217 \\
    \midrule

    \multirow{3}{*}{\rotatebox{90}{\textbf{XAI}}}
      & None              &  264 & \textbf{920}\tiny$\pm$1227   & \textbf{62.2}\tiny$\pm$116 \\
      & SHAP              &  198 & 952\tiny$\pm$1182            & 65.8\tiny$\pm$130 \\
      & LIME              &  198 & 1601\tiny$\pm$3280           & 79.4\tiny$\pm$145 \\
    \midrule

    \multirow{3}{*}{\rotatebox{90}{\textbf{LLM}}}
      & GPT-4o            &  240 & \textbf{575}\tiny$\pm$667    & \textbf{11.7}\tiny$\pm$17.6 \\
      & Llama-3           &  240 & 738\tiny$\pm$846             & 151.2\tiny$\pm$182 \\
      & DeepSeek-R1       &  180 & 2407\tiny$\pm$3482           & 33.7\tiny$\pm$49.5 \\
    \midrule

    \multirow{4}{*}{\rotatebox{90}{\textbf{Model}}}
      & SARIMAX           &   66 & \textbf{1053}\tiny$\pm$1458  & 83.1\tiny$\pm$148 \\
      & RandomForest      &  198 & 1084\tiny$\pm$2041           & \textbf{64.2}\tiny$\pm$123 \\
      & XGBoost           &  198 & 1119\tiny$\pm$1956           & 67.2\tiny$\pm$130 \\
      & MLP               &  198 & 1225\tiny$\pm$2402           & 69.0\tiny$\pm$129 \\
    \bottomrule
  \end{tabular}
  \caption{Generation efficiency by experimental factor (mean $\pm$ SD), ordered by tokens. \textbf{Bold} = best per factor. N varies by design: CoT excludes DeepSeek-R1; SARIMAX uses only ``none'' XAI.}
  \label{tab:efficiency}
\end{table}

\section{Experiments and Results}
\label{sec:results}

For each research question, we test overall effects using \textbf{one-way ANOVA}. Beyond $p$-values, we report \textbf{omega-squared ($\omega^2$)} \citep{kroes2023demystifying} as a measure of effect size -- the proportion of variance in scores attributable to the factor (e.g., $\omega^2=.06$ for XAI means 6\% of score variation is due to XAI choice). Following \citet{field2013discovering}, we interpret: $<.01$ negligible, .01--.06 small, .06--.14 medium, $>.14$ large. Where ANOVA is significant, we conduct \textbf{pairwise $t$-tests} reporting \textbf{Cohen's $d$} \citep{cohen1988statistical} -- the standardized difference between group means. We interpret: $<.2$ negligible, .2--.5 small, .5--.8 medium, $>.8$ large.

Because we test five evaluation dimensions simultaneously, we apply \textbf{Benjamini-Hochberg FDR correction} \citep{benjamini1995controlling} to control false discovery rates, reporting both raw $p$ and corrected $p_{\text{FDR}}$ values. We also fit \textbf{factorial models} \citep{maxwell2017designing} controlling for other experimental factors (e.g., testing XAI effect while controlling for LLM and Strategy), and test for \textbf{interactions} to check whether effects vary across conditions.

Below we present the key findings; full statistical tables, robustness checks (Welch ANOVA, Games-Howell tests), and interaction analyses are in Appendix~\ref{app:statistical}.

\subsection{RQ1: Does XAI Improve NLE Quality?}

\vspace{-3mm}
\begin{table}[H]
\centering
\small
\begin{tabular}{lccccc}
\toprule
\textbf{XAI} & \textbf{Acc.} & \textbf{Lay} & \textbf{Exp.} & \textbf{Help.} & \textbf{Avg.} \\
\midrule
LIME & \textbf{4.57} & 4.31 & 3.94 & 4.31 & \textbf{4.28} \\
SHAP & 4.53 & \textbf{4.32} & \textbf{3.97} & 4.30 & 4.28 \\
None & 4.49 & 4.31 & 3.78 & \textbf{4.38} & 4.24 \\
\bottomrule
\end{tabular}
\caption{G-Eval scores by XAI condition, sorted by average (SARIMAX excluded). Acc./Lay/Exp./Help.\ are the four evaluation dimensions from \S\ref{sec:rubrics}; values are means across all samples per condition; Avg.\ is the mean of the four dimensions. Same format for all results tables. \textbf{Bold} = best per column.}
\label{tab:xai_results}
\end{table}

One-way ANOVA reveals a significant XAI effect only for Expert Relevancy ($\omega^2=.02$, small effect, survives FDR); other dimensions show no significant differences. SHAP and LIME both outperform no-XAI ($d\approx0.3$) but do not differ from each other. Significant XAI$\times$LLM interactions indicate the benefit varies by LLM: GPT-4o gains $+0.3$ points with XAI, DeepSeek-R1 gains $+0.2$ points, while Llama-3 shows no improvement.

\textbf{Finding 1:} XAI improves only Expert Relevancy (small effect, $d\approx0.3$); other dimensions show no benefit. SHAP and LIME perform equivalently.

\subsection{RQ2: Which LLM Produces the Best NLEs?}

\vspace{-2mm}
\begin{table}[H]
\centering
\small
\begin{tabular}{lccccc}
\toprule
\textbf{LLM} & \textbf{Acc.} & \textbf{Lay} & \textbf{Exp.} & \textbf{Help.} & \textbf{Avg.} \\
\midrule
DeepSeek & 4.71 & \textbf{4.72} & \textbf{4.45} & \textbf{4.78} & \textbf{4.66} \\
GPT-4o & \textbf{4.73} & 4.20 & 3.95 & 4.28 & 4.29 \\
Llama-3 & 4.18 & 4.05 & 3.39 & 3.99 & 3.90 \\
\bottomrule
\end{tabular}
\caption{G-Eval scores by generator LLM, sorted by average. \textbf{Bold} = best per column.}
\label{tab:llm_results}
\end{table}

LLM choice shows the strongest effect across all factors. One-way ANOVA reveals significant effects on all four dimensions with large effect sizes: Accuracy ($\omega^2=.20$), Lay Relevancy ($\omega^2=.25$), Expert Relevancy ($\omega^2=.50$), and Helpfulness ($\omega^2=.38$); all survive FDR correction. DeepSeek-R1 outperforms GPT-4o and Llama-3 on three dimensions with large effects ($d=1.2$--$2.5$). On Accuracy, DeepSeek and GPT-4o are equivalent, while both outperform Llama-3 ($d\approx0.9$). Effects persist when controlling for XAI and Strategy. However, quality comes at a cost (Table~\ref{tab:efficiency}): DeepSeek-R1 generates 4$\times$ more tokens than GPT-4o.

\textbf{Finding 2:} LLM choice has the largest effect on NLE quality ($\omega^2$ up to .50). DeepSeek-R1 produces the best explanations but at 4$\times$ the token cost.

\subsection{RQ3: Does ML Model Quality Affect NLEs?}

\vspace{-2mm}
\begin{table}[H]
\centering
\small
\begin{tabular}{lccccc}
\toprule
\textbf{Model} & \textbf{Acc.} & \textbf{Lay} & \textbf{Exp.} & \textbf{Help.} & \textbf{Avg.} \\
\midrule
XGBoost & \textbf{4.56} & \textbf{4.53} & \textbf{4.00} & \textbf{4.43} & \textbf{4.38} \\
RF & 4.53 & 4.29 & 3.82 & 4.27 & 4.23 \\
MLP & 4.50 & 4.12 & 3.86 & 4.29 & 4.19 \\
\bottomrule
\end{tabular}
\caption{G-Eval scores by ML model, sorted by average. XGBoost (highest R$^2$) yields best NLEs. \textbf{Bold} = best per column.}
\label{tab:model_results}
\end{table}

ML Model choice (SARIMAX analyzed separately in RQ4) significantly affects three dimensions: Lay Relevancy ($\omega^2=.12$, medium), Helpfulness ($\omega^2=.02$, small), and Expert Relevancy ($\omega^2=.02$, small); all survive FDR. Accuracy shows no model effect ($p=.55$). XGBoost (highest R$^2$) outperforms both RF and MLP on all three significant dimensions: large effect on Lay Relevancy ($d=0.48$--$0.82$), small effects on Expert Relevancy and Helpfulness ($d=0.24$--$0.33$). Effects persist when controlling for LLM, XAI, and Strategy. No Model$\times$LLM or Model$\times$Strategy interactions emerge (all $p>.15$).

\textbf{Finding 3:} Better-performing ML models yield better NLEs (medium effect on Lay Relevancy, small on Expert/Helpfulness). Accuracy is unaffected by model quality.

\subsection{RQ4: Does Inherent Interpretability Help?}

\vspace{-3mm}
\begin{table}[H]
\centering
\small
\begin{tabular}{lccccc}
\toprule
\textbf{Model} & \textbf{Acc.} & \textbf{Lay} & \textbf{Exp.} & \textbf{Help.} & \textbf{Avg.} \\
\midrule
XGBoost & \textbf{4.51} & \textbf{4.61} & \textbf{3.89} & \textbf{4.54} & \textbf{4.39} \\
RF & 4.46 & 4.26 & 3.68 & 4.29 & 4.17 \\
MLP & 4.50 & 4.07 & 3.76 & 4.32 & 4.16 \\
SARIMAX & 4.47 & 4.04 & 3.76 & 4.12 & 4.10 \\
\bottomrule
\end{tabular}
\caption{NLE quality at XAI=``none'' ($N=66$ each), sorted by average. \textbf{Bold} = best per column.}
\label{tab:sarimax_results}
\end{table}

SARIMAX offers inherently interpretable coefficients (AR, MA, seasonal terms), unlike black-box ML models. Table~\ref{tab:sarimax_results} compares models at XAI=``none'' (fair since SARIMAX cannot use SHAP/LIME).

SARIMAX yields the lowest scores despite having the second-highest R$^2$. Significant Model effects emerge on Lay Relevancy ($\omega^2=.17$, medium) and Helpfulness ($\omega^2=.08$, small); both survive FDR. SARIMAX significantly underperforms XGBoost with large effects ($d=-0.98$ to $-1.16$) and even underperforms lower-R$^2$ models RF and MLP ($d=-0.37$ to $-0.44$).

\textbf{Finding 4:} In our setting, SARIMAX yielded lower NLE quality than ML models despite higher prediction accuracy.

\subsection{RQ5: Prompting Strategy Comparison}

\vspace{-2mm}
\begin{table}[H]
\centering
\small
\begin{tabular}{lccccc}
\toprule
\textbf{Strategy} & \textbf{Acc.} & \textbf{Lay} & \textbf{Exp.} & \textbf{Help.} & \textbf{Avg.} \\
\midrule
Self-cons. & 4.72 & \textbf{4.57} & \textbf{4.05} & 4.43 & \textbf{4.44} \\
Zero-shot & 4.70 & 4.36 & 4.00 & 4.36 & 4.35 \\
Reflexion & \textbf{4.74} & 4.39 & 3.96 & 4.31 & 4.35 \\
Role-based & 4.58 & 4.48 & 3.84 & 4.36 & 4.32 \\
Few-shot & 4.33 & 4.20 & 3.94 & \textbf{4.53} & 4.25 \\
CoT Zero & 4.71 & 4.19 & 3.64 & 4.12 & 4.17 \\
Meta & 4.22 & 3.97 & 3.80 & 4.06 & 4.01 \\
CoT Few & 4.13 & 4.02 & 3.68 & 4.22 & 4.01 \\
\bottomrule
\end{tabular}
\caption{G-Eval scores by prompting strategy, sorted by average.}
\label{tab:strategy_results}
\end{table}

Strategy effects are significant on all four dimensions with medium-to-large effect sizes: Accuracy ($\omega^2=.14$), Lay Relevancy ($\omega^2=.12$), Helpfulness ($\omega^2=.08$), and Expert Relevancy ($\omega^2=.04$); all survive FDR. Self-consistency ranks highest, followed by zero-shot and reflexion. Simple zero-shot outperforms complex approaches like chain-of-thought and meta-prompting. Strong Strategy$\times$LLM interactions ($\omega^2=.28$--$.41$) indicate optimal strategy depends on the generator LLM.

Self-consistency consumes 7$\times$ more tokens than zero-shot and requires 10$\times$ longer generation time for only 0.09 points improvement (Table~\ref{tab:efficiency}).

\textbf{Finding 5:} Self-consistency achieves highest quality but at 7$\times$ the cost of zero-shot. Zero-shot offers near-equivalent quality; chain-of-thought underperforms.


\section{Discussion}

\subsection{What the Evaluation Dimensions Reveal}

The four G-Eval dimensions capture distinct aspects of NLE quality. Our results show they respond differently to experimental factors, but LLM capability dominates all of them.

\textbf{Accuracy} measures factual correctness in representing provided information (numerical values, feature names, metrics). It is primarily determined by LLM choice ($\omega^2=.20$) and prompting strategy ($\omega^2=.14$). XAI condition has no effect: LLMs accurately report whatever information they receive, whether that includes feature attributions or not.

\textbf{Lay Relevancy} captures whether non-experts can understand the prediction rationale, assess trustworthiness, and derive actionable insights. It shows no XAI benefit but responds to ML model quality ($\omega^2=.12$) and prompting strategy ($\omega^2=.12$). The absent XAI effect aligns with research showing that lay users struggle to interpret feature attributions \citep{salih2025perspective, alufaisan2020explainable}. Better-performing models provide richer context for explanations; how the LLM frames that context matters equally.

\textbf{Expert Relevancy} assesses technical depth and whether ML practitioners can assess model behavior. It is overwhelmingly determined by LLM capability ($\omega^2=.50$). XAI shows a statistically significant but small effect ($\omega^2=.02$): including feature attributions provides marginally more technical content, but the LLM's ability to present information coherently explains 25$\times$ more variance.

\textbf{Helpfulness} evaluates whether users can judge prediction accuracy from cues like uncertainty ranges or error context. It shows the second-largest LLM effect ($\omega^2=.38$) with no XAI benefit. Communicating prediction reliability depends on LLM capability, not on whether XAI outputs are included.

\subsection{SHAP vs. LIME}

SHAP and LIME show no significant differences across any dimension. While these methods differ algorithmically (SHAP uses Shapley values from game theory, LIME fits local surrogate models), both produce similar feature attribution outputs for tabular data \citep{salih2025perspective}. The bottleneck for NLE quality appears to be the LLM's translation of attributions into narrative, not the XAI method itself.

The significant XAI$\times$LLM interactions reveal that the benefit of including XAI outputs depends on the generator. For Expert Relevancy, GPT-4o gains $+0.3$ points with XAI (from 3.79 to 4.13), DeepSeek-R1 gains $+0.2$ points, while Llama-3 shows no significant improvement. This suggests that weaker models may not effectively leverage XAI information even when provided.

\subsection{The Interpretability Paradox}

SARIMAX yields lower NLE quality than ML models despite higher prediction accuracy (R$^2$=.55 vs.\ .21--.37 for MLP/RF) and being inherently interpretable through its statistical coefficients. This is paradoxical: econometricians have interpreted classical time-series models for decades, and the very existence of XAI methods reflects that ML models are \emph{not} inherently interpretable to humans. Yet LLMs produce better explanations for the opaque models than for the transparent ones.

We can only hypothesize about the cause, and it would need to be tested. One possibility is that LLM training data plays a role: feature-importance language (``temaperature contributed +0.3'') may be more common than statistical coefficient notation (AR, MA, seasonal terms). It may also be easier to translate XAI outputs into natural language, while AR, MA, and seasonal coefficients may be harder to explain accessibly, especially for lay users. If this is indeed an LLM limitation rather than a property of model types, fine-tuning on statistical content or improved prompting may close this gap.

\subsection{Strategy Selection and Cost-Effectiveness}

The strong Strategy$\times$LLM interactions ($\omega^2=.28$--$.41$) indicate that optimal prompting strategy varies substantially by generator. No single strategy dominates across all LLMs.

Self-consistency achieves the highest overall scores (4.21 average) but at 7$\times$ the token cost of zero-shot, which scores 4.12. This amounts to a gain of 0.09 points per 7$\times$ increase in computational cost. Whether this trade-off is worthwhile depends on the deployment context: for high-stakes medical or financial explanations, marginal quality improvements may justify the cost; for high-volume automated reporting, zero-shot offers near-equivalent quality at dramatically lower cost.

Chain-of-thought's underperformance is surprising given its success on other tasks. One possibility is that explicit reasoning chains add verbosity without improving the final explanation, or that CoT is better suited to tasks requiring multi-step logical inference. Regardless of the mechanism, our results suggest CoT benefits are task-dependent, not universal.

\section{Conclusion}

We presented a systematic factorial study of factors affecting LLM-generated natural language explanations from XAI outputs. Across 4 ML models, 3 XAI conditions, 3 LLMs, and 8 prompting strategies, we generated and evaluated 660 explanations using G-Eval with dual LLM judges.

Our key findings: (1) XAI provides modest value: SHAP/LIME improve only Expert Relevancy ($\omega^2=.02$), while LLMs generate coherent explanations from context alone. (2) LLM choice dominates all other factors ($\omega^2$ up to .50), with DeepSeek-R1 outperforming GPT-4o and Llama-3. (3) We observe an \emph{interpretability paradox}: in our setting, SARIMAX yielded lower NLE quality than ML models despite higher prediction accuracy. (4) Zero-shot prompting is competitive with self-consistency at 7$\times$ lower cost; chain-of-thought underperforms.

These findings challenge the assumption that sophisticated XAI methods automatically yield better explanations. The LLM's capability to translate technical outputs matters more than the XAI method producing them. We offer testable hypotheses for the interpretability paradox.

\clearpage
\section*{Limitations}

\textbf{Scope.} Our experiments focus on a single domain (household energy forecasting) using data from one household. Energy consumption is relatively intuitive -- features like temperature and time-of-week have common-sense interpretations. Domains requiring specialized knowledge (e.g., genomics, materials science) or different data modalities (e.g., images, text) may yield different patterns.

\textbf{Methods.} We test only two XAI methods (SHAP and LIME), three LLMs (GPT-4o, DeepSeek-R1, Llama-3), and one classical model (SARIMAX). Other XAI approaches (e.g., attention-based, counterfactual, concept-based), other LLMs (e.g., Claude, Gemini), and other classical models (e.g., VAR, exponential smoothing) may behave differently. The interpretability paradox warrants replication with additional classical models. Furthermore, we use only prompt-based generation; fine-tuning or retrieval-augmented approaches may yield different results. Results may also be sensitive to prompt formulation.

\textbf{Evaluation.} We rely on LLM-based evaluation (G-Eval) rather than human annotators. Despite our bias mitigation strategies, LLM judges may have systematic biases not captured by our design. More importantly, we measure explanation \emph{quality} but not \emph{utility}. Research shows that explanation quality and user performance are often weakly correlated \citep{schemmer2022should}: users may over-rely on confident-sounding explanations or ignore technically accurate but complex ones. DeepSeek-R1 produces higher-quality explanations than GPT-4o, but whether users make better decisions with them remains untested. Similarly, the interpretability paradox concerns NLE quality, not whether users actually understand SARIMAX better when explained by humans versus LLMs.

\textbf{Generalizability.} Our findings may not generalize beyond time-series forecasting, as other domains (e.g., image classification, NLP) involve different XAI methods and explanation structures. The rapid pace of LLM development also means our specific rankings may not persist.
\bibliography{custom}


\appendix

\section{Experimental Details}
\label{app:experimental}

This section provides implementation details for data preprocessing, model training, and XAI computation.

\subsection{Data Preprocessing}

We use the UCI Individual Household Electric Power Consumption dataset \citep{hebrail2012individual}, containing 2,075,259 minute-level measurements from a household in Sceaux, France (near Paris), recorded from December 2006 to November 2010. The preprocessing pipeline: (1) handles missing values ($\approx$1.25\%) via linear interpolation, (2) converts kilowatt-minutes to kilowatt-hours, and (3) aggregates to weekly totals anchored on Mondays (ISO week alignment):

\begin{tcolorbox}[codebox, title={Weekly Aggregation}]
\begin{lstlisting}[style=pythonstyle]
def load_weekly(file_path, country="FR"):
    df = pd.read_csv(file_path, sep=";", na_values="?")
    df["DateTime"] = pd.to_datetime(df["Date"] + " " + df["Time"])

    # Handle missing values + convert kW-min -> kWh
    df["Global_active_power"] = (
        df["Global_active_power"].astype(float).interpolate() * (1/60)
    )

    # Weekly aggregation anchored on Monday (ISO alignment)
    w = df["Global_active_power"].resample("W-MON").sum()
    w = w.to_frame("True_Value").reset_index()

    # ISO week number (1-52)
    w["weekofyear"] = w["DateTime"].dt.isocalendar().week

    # Count holidays in each week
    fr_holidays = holidays.CountryHoliday(country)
    w["holiday_week_count"] = w["DateTime"].apply(
        lambda d: sum((d - pd.Timedelta(days=i)).date() in fr_holidays
                      for i in range(7))
    )
    return w
\end{lstlisting}
\end{tcolorbox}

\noindent\textbf{Calendar features:} \texttt{weekofyear} captures the ISO week number (1--52) for annual seasonality patterns (e.g., higher winter consumption). \texttt{holiday\_week\_count} counts French public holidays within each week using the \texttt{holidays} Python package---for each week ending on Monday, we count holidays in the preceding 7-day window.

\noindent\textbf{Lag features:} Seven autoregressive features (\texttt{lag\_1} through \texttt{lag\_7}) capture the previous seven weeks of consumption:

\begin{tcolorbox}[codebox, title={Feature Engineering}]
\begin{lstlisting}[style=pythonstyle]
N_LAGS = 7
for i in range(1, N_LAGS + 1):
    weekly[f"lag_{i}"] = weekly["True_Value"].shift(i)
weekly = weekly.dropna()

FEATURES = [f"lag_{i}" for i in range(1, N_LAGS+1)] + \
           ["weekofyear", "holiday_week_count"]

# Chronological 70/30 split
cut = int(len(weekly) * 0.70)
train, test = weekly.iloc[:cut], weekly.iloc[cut:]
\end{lstlisting}
\end{tcolorbox}

\subsection{Model Configurations}

\paragraph{XGBoost.} Hyperparameters selected via 200-iteration random search with 4-fold time-series cross-validation:

\begin{tcolorbox}[codebox, title={XGBoost Configuration}]
\begin{lstlisting}[style=pythonstyle]
XGBRegressor(
    n_estimators     = 1190,
    learning_rate    = 0.00497,
    max_depth        = 5,
    subsample        = 0.805,
    colsample_bytree = 0.725,
    min_child_weight = 6,
    gamma            = 0.248,
    reg_lambda       = 0.530,
    early_stopping_rounds = 200
)
\end{lstlisting}
\end{tcolorbox}

\paragraph{SARIMAX.} Orders selected via stepwise AICc search using \texttt{pmdarima.auto\_arima} with: non-seasonal AR/MA orders $p,q \in [0,3]$, seasonal AR/MA orders $P,Q \in [0,2]$, forced seasonal differencing $D=1$, automatic non-seasonal differencing, Canova--Hansen seasonality test, and annual seasonality $m=52$. The selected configuration:

\begin{tcolorbox}[codebox, title={SARIMAX Configuration}]
\begin{lstlisting}[style=pythonstyle]
SARIMAX(
    endog = y_train,
    exog  = train[["holiday_week_count"]],
    order = (2, 0, 1),
    seasonal_order = (1, 1, 0, 52),
    enforce_stationarity = False,
    enforce_invertibility = False,
    trend = 'c'
)
\end{lstlisting}
\end{tcolorbox}

\noindent For test-set evaluation, we use rolling one-step-ahead forecasts: after each prediction, the true observed value is appended to the model state (without refitting) before forecasting the next week.

\paragraph{RF \& MLP.} Deliberately weak configurations to span model quality:

\begin{tcolorbox}[codebox, title={Weak Model Configurations}]
\begin{lstlisting}[style=pythonstyle]
# Random Forest: shallow stumps
RandomForestRegressor(
    n_estimators=10, max_depth=1,
    min_samples_leaf=25, max_features=0.08
)

# MLP: minimal capacity
make_pipeline(
    StandardScaler(),
    MLPRegressor(
        hidden_layer_sizes=(8,), activation="tanh",
        solver="sgd", learning_rate_init=0.30,
        max_iter=30, n_iter_no_change=2
    )
)
\end{lstlisting}
\end{tcolorbox}

\subsection{XAI Computation}

\paragraph{SHAP.} TreeExplainer for tree-based models; KernelExplainer with 100-sample background for MLP:

\begin{tcolorbox}[codebox, title={SHAP Computation}]
\begin{lstlisting}[style=pythonstyle]
# Tree models (XGBoost, RandomForest)
explainer = shap.TreeExplainer(model)
shap_values = explainer.shap_values(X_test)
base_value = explainer.expected_value

# MLP (KernelExplainer)
background = shap.sample(X_train, 100, random_state=42)
explainer = shap.KernelExplainer(model.predict, background)
shap_values = explainer.shap_values(X_test)
\end{lstlisting}
\end{tcolorbox}

\paragraph{LIME.} Local surrogate with standardized features:

\begin{tcolorbox}[codebox, title={LIME Configuration}]
\begin{lstlisting}[style=pythonstyle]
explainer = LimeTabularExplainer(
    training_data = scaler.transform(X_train),
    feature_names = FEATURES,
    mode = "regression",
    discretize_continuous = False,
    sample_around_instance = True,
    kernel_width = 3,
    random_state = 42
)
# 8000 samples per explanation
exp = explainer.explain_instance(row, predict_fn,
                                  num_samples=8000)
\end{lstlisting}
\end{tcolorbox}

\subsection{Prediction Visualization}

\begin{figure*}[t]
    \centering
    \includegraphics[width=\textwidth]{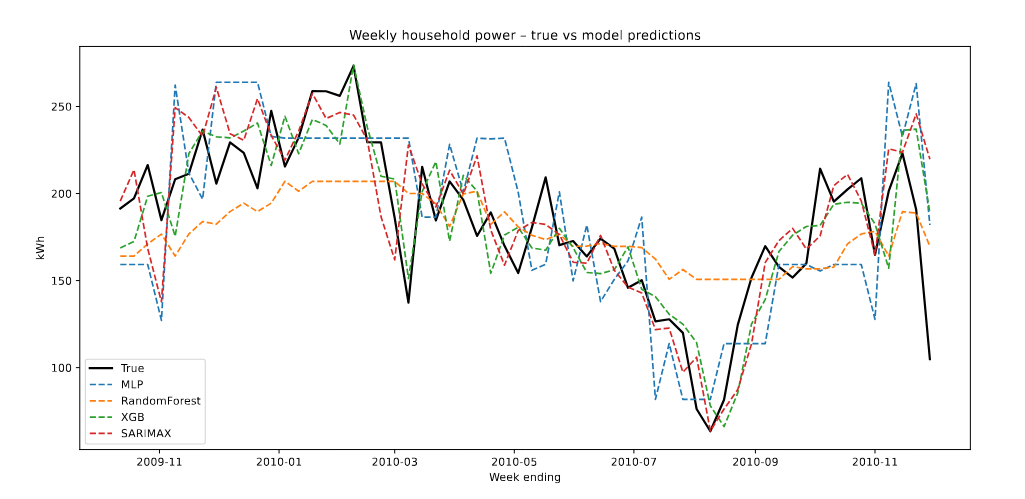}
    \caption{Test-set predictions vs.\ actual weekly household energy consumption (kWh). XGBoost (R$^2$=0.69) captures temporal patterns effectively; Random Forest (R$^2$=0.37) shows moderate tracking; SARIMAX (R$^2$=0.55) handles seasonality but misses peaks; MLP (R$^2$=0.21) exhibits high variance with poor trend-following.}
    \label{fig:model_preds_large}
\end{figure*}

Figure~\ref{fig:model_preds_large} shows test-set predictions for all four models against actual weekly consumption. XGBoost closely tracks the true series, while MLP exhibits high variance and poor trend-following.

\subsection{Test Instance Details}
\label{app:test_instances}

We select three test weeks using tercile-based sampling from the 52-week test set. The selection criterion requires absolute prediction errors to rank as: $|\text{XGB}| < |\text{SARIMAX}| < |\text{RF}| < |\text{MLP}|$, ensuring each instance demonstrates the quality gradient across models.

\begin{table}[H]
\centering
\small
\begin{tabular}{llrrrr}
\toprule
\textbf{Tercile} & \textbf{Date} & \textbf{Week} & \textbf{True (kWh)} \\
\midrule
Early  & 2009-11-09 & 46 & 208.20 \\
Middle & 2010-04-26 & 17 & 169.98 \\
Late   & 2010-07-26 & 30 & 119.95 \\
\bottomrule
\end{tabular}
\caption{Selected test instances.}
\label{tab:test_instances}
\end{table}

\begin{table}[H]
\centering
\small
\begin{tabular}{llrr}
\toprule
\textbf{Date} & \textbf{Model} & \textbf{Pred.} & \textbf{Err. (\%)} \\
\midrule
\multirow{4}{*}{2009-11-09}
  & XGBoost & 175.4 & $-$15.7 \\
  & SARIMAX & 249.3 & +19.8 \\
  & RF & 164.1 & $-$21.2 \\
  & MLP & 262.3 & +26.0 \\
\midrule
\multirow{4}{*}{2010-04-26}
  & XGBoost & 176.1 & +3.6 \\
  & SARIMAX & 158.8 & $-$6.6 \\
  & RF & 189.5 & +11.5 \\
  & MLP & 231.8 & +36.4 \\
\midrule
\multirow{4}{*}{2010-07-26}
  & XGBoost & 125.0 & +4.2 \\
  & SARIMAX & 97.3 & $-$18.9 \\
  & RF & 156.3 & +30.3 \\
  & MLP & 81.8 & $-$31.8 \\
\bottomrule
\end{tabular}
\caption{Predictions and errors per model.}
\label{tab:instance_errors}
\end{table}

\begin{table}[H]
\centering
\small
\begin{tabular}{lrrr}
\toprule
\textbf{Feature} & \textbf{Nov 9} & \textbf{Apr 26} & \textbf{Jul 26} \\
\midrule
lag\_1 & 184.7 & 189.2 & 127.8 \\
lag\_2 & 216.4 & 175.6 & 126.6 \\
lag\_3 & 197.2 & 196.4 & 150.3 \\
lag\_4 & 191.4 & 207.0 & 145.9 \\
lag\_5 & 166.3 & 184.6 & 168.4 \\
lag\_6 & 162.3 & 215.4 & 174.2 \\
lag\_7 & 189.8 & 137.4 & 163.9 \\
weekofyear & 46 & 17 & 30 \\
holiday\_count & 0 & 0 & 0 \\
\bottomrule
\end{tabular}
\caption{Feature values (lags in kWh).}
\label{tab:instance_features}
\end{table}

\section{Prompt Templates}
\label{app:prompts}

All prompting strategies share a common two-message structure: a \textbf{System Message} establishing the task and output constraints, and a \textbf{Human Message} providing the specific forecasting context.

\subsection{Prompt Structure}

\begin{tcolorbox}[systemprompt, title=System Message (Common)]
Interpret the time-series forecasting context that follows.\\[4pt]
Your goal is to help a non-technical lay user. The user does not have a background in statistical models, machine learning, times-series or explainability methods (e.g. SHAP, LIME). The user needs to understand:\\[2pt]
\hspace*{1em}-- \textbf{why} the model produced this forecast, and\\
\hspace*{1em}-- \textbf{how much confidence} can be placed in it.\\[4pt]
OUTPUT RULES\\
\hspace*{1em}-- Write \textbf{up to six bullet points}  -- no more.\\
\hspace*{1em}-- Keep the whole response \textbf{<= 200 words}.\\
\hspace*{1em}-- Use plain language, but key time-series terms such as lag, trend, seasonality, baseline, error are fine.\\
\hspace*{1em}-- Do not reveal code.
\end{tcolorbox}

\begin{tcolorbox}[humanprompt, title=Human Message (Common Structure)]
The following is about time series data with a single-step ahead prediction, where the model predicts the next value in the time series based on previous observations.\\[4pt]
\textbf{Data Domain:} Energy Consumption\\[2pt]
\textbf{Dataset Description:}\\
\hspace*{1em}-- The dataset contains 2,075,259 measurements from a house in Sceaux, France (near Paris), Dec 2006--Nov 2010.\\
\hspace*{1em}-- Recorded at one-minute resolution, then resampled weekly.\\
\hspace*{1em}-- Lag\_1..Lag\_7, ISO week number and number of public holidays per week were added as features.\\
\hspace*{1em}-- Target: weekly global active power (kWh).\\[2pt]
\textbf{Model Used:} [XGBoost / RandomForest / MLP / SARIMAX]\\
\textbf{Model Performance:}\\
\hspace*{1em}-- MAE: [value]\\
\hspace*{1em}-- RMSE: [value]\\
\hspace*{1em}-- R\textsuperscript{2}: [value]\\[2pt]
\textbf{Prediction:} [predicted value in kWh]\\[2pt]
\textbf{Instance Features or Context:}\\
\hspace*{1em}lag\_1: [value], lag\_2: [value], ..., lag\_7: [value]\\
\hspace*{1em}weekofyear: [1--52], holiday\_week\_count: [0--N]\\[2pt]
\textbf{[SHAP/LIME] values:} (sorted by $|$magnitude$|$, omitted for XAI=none)\\
\hspace*{1em}[feature]: [+/-value]\\
\hspace*{1em}...\\
\hspace*{1em}The expected/base value for [SHAP/LIME]: [baseline]\\[2pt]
\textit{(For LIME only:)} Each LIME value is already $\beta_i \times x_i$: the additive kWh contribution of that feature for this instance.
\end{tcolorbox}

\noindent\textbf{SARIMAX-specific addition:} For SARIMAX prompts, the human message additionally appends the full \texttt{statsmodels} model summary (coefficients, standard errors, diagnostics) followed by a note:

\begin{tcolorbox}[promptbox]
{[}Note{]} The SARIMAX order/seasonal\_order were chosen via an auto\_arima stepwise AICc search. Lagged demand effects are captured implicitly by the AR and MA terms; only `holiday\_week\_count' enters as an explicit regressor.
\end{tcolorbox}

\subsection{Strategy-Specific Additions}

\paragraph{Zero-shot.} Uses only the system prompt above with no additional instructions.

\paragraph{Few-shot.}
\label{app:fewshot}
Prepends four complete Human/Assistant example pairs demonstrating different domains and XAI methods. Below is one example; the remaining three are available in our GitHub repository.

\begin{tcolorbox}[humanprompt, title=Human]
\textbf{Data Domain:} Retail Sales\\[2pt]
\textbf{Dataset Description:}\\
\hspace*{1em}-- 469 weekly sales totals from a mid-size supermarket chain in Madrid, Jan 2015 -- Dec 2023\\
\hspace*{1em}-- Originally daily cash-register data, aggregated to weekly sums\\
\hspace*{1em}-- Features: Lag\_1...Lag\_8, ISO week number, holidays per week, Promotion\_Intensity (0--1)\\
\hspace*{1em}-- Target: weekly gross sales (thousand EUR)\\[2pt]
\textbf{Model Used:} XGBRegressor\\
\textbf{Model Performance:} MAE: 12.3, RMSE: 16.8, R2: 0.874\\[2pt]
\textbf{Prediction:} 242.70\\[2pt]
\textbf{Instance Features:} lag\_1: 238.4, lag\_2: 245.1, ..., weekofyear: 48, holiday\_week\_count: 1, promotion\_intensity: 0.65\\[2pt]
\textbf{SHAP values:} promotion\_intensity: +6.9, lag\_1: +4.2, holiday\_week\_count: +3.5, lag\_2: -2.8, lag\_3: -1.1, ...\\
\textbf{Base value:} 235.1
\end{tcolorbox}

\begin{tcolorbox}[assistantprompt, title=Assistant]
-- SHAP starts from a baseline of \textasciitilde235k EUR and shows how each feature nudges the forecast up (positive) or down (negative).\\
-- Promotion intensity (0.65) is the main driver, adding \textasciitilde7k EUR.\\
-- Last-week sales (lag\_1 = 238k EUR) add \textasciitilde4k EUR, showing continued momentum.\\
-- One public holiday lifts demand by \textasciitilde3.5k EUR, but the boost is modest.\\
-- Week 48 sits before Christmas; the tiny +0.4k EUR seasonal bump hints demand hasn't peaked yet.\\
-- Negative pulls from lags 2-4 temper the forecast; 242.7k EUR is plausible within the +/-16.8k EUR RMSE.
\end{tcolorbox}

\paragraph{CoT Zero-shot.} Appends to system prompt:
\begin{tcolorbox}[promptbox]
Think step-by-step.
\end{tcolorbox}

\paragraph{CoT Few-shot.} Combines few-shot examples with CoT instruction.

\paragraph{Role-based.} Appends to system prompt:
\begin{tcolorbox}[promptbox]
You are a seasoned data-scientist and approachable teacher.\\
You love turning complex numbers into everyday language so\\
home-owners can make smart, confident decisions.
\end{tcolorbox}

\paragraph{Meta-prompting.} Uses a structured 6-step reasoning template prepended to the human message:
\begin{tcolorbox}[promptbox]
Let's think step by step:\\
\hspace*{1em}-- Context: Summarise the problem domain and briefly name the XAI method.\\
\hspace*{1em}-- Data Insights: Note dataset size, key features, possible biases.\\
\hspace*{1em}-- Model/Parameters: Mention any parameter that affects performance.\\
\hspace*{1em}-- Performance Metrics: Explain MAE, RMSE, R2.\\
\hspace*{1em}-- XAI / Feature Importance: (1) Top factors. (2) Why they matter. (3) Surprises.\\
\hspace*{1em}-- Summary: Tie the insights into a final answer.
\end{tcolorbox}

\paragraph{Reflexion.} Iterative self-improvement loop with up to 6 trials. Each iteration: (1) generate draft, (2) grade via GPT-4o, (3) self-reflect. Stops early if all scores $\geq$4. Memory retains last 3 reflections.

\begin{tcolorbox}[promptbox, title={Grading Prompt (GPT-4o)}]
You are evaluating a time-series forecast explanation for a lay reader.\\[4pt]
\textbf{Ground-truth facts}\\
Prediction: [value]\\
Top features $\pm$ values: [XAI attributions]\\[4pt]
\textbf{Explanation to grade}\\
$<<<$ [draft explanation] $>>>$\\[4pt]
Rate each criterion 1--5 (briefly justify $\leq$15 words):\\
\hspace*{1em}1. Accuracy / Faithfulness\\
\hspace*{1em}2. Coverage of key drivers\\
\hspace*{1em}3. Confidence communication\\
\hspace*{1em}4. Clarity for a lay reader\\[4pt]
Respond with one valid JSON line:\\
\texttt{\{"accuracy": x, "coverage": y, "confidence": z, "clarity": w, "justifications": \{...\}\}}
\end{tcolorbox}

\begin{tcolorbox}[promptbox, title={Reflection Prompt}]
Outcome: [SUCCESS/FAIL].\\
Give ONE short sentence advising yourself how to improve accuracy, coverage, confidence, clarity next time.
\end{tcolorbox}

\paragraph{Self-consistency.} Generates 7 independent drafts at temperature $\tau$=1.0, then synthesizes into one cohesive explanation at $\tau$=0.2.

\begin{tcolorbox}[promptbox, title={Synthesis Prompt ($\tau$=0.2)}]
Below are multiple candidate explanations produced by the model for the same input. Please create a single, cohesive explanation that:\\
\hspace*{1em}-- Retains the most insightful sentences from each.\\
\hspace*{1em}-- Resolves contradictions.\\
\hspace*{1em}-- Still obeys the format instructions.\\[4pt]
=== CANDIDATES ===\\
--- Candidate \#1 ---\\
{[}draft 1{]}\\
--- Candidate \#2 ---\\
{[}draft 2{]}\\
... (7 total)
\end{tcolorbox}

\section{Example Input and Output}
\label{app:examples}

Below is a complete example from our energy consumption domain using zero-shot prompting with SHAP.

\subsection{Input Context}

\begin{tcolorbox}[humanprompt, title={Input (Zero-shot with SHAP)}]
\textbf{Data Domain:} Household Energy Consumption\\[2pt]
\textbf{Dataset Description:}\\
\hspace*{1em}-- 200 weekly power consumption totals from a French household, Dec 2006 -- Nov 2010\\
\hspace*{1em}-- Originally minute-level data, aggregated to weekly sums (kWh)\\
\hspace*{1em}-- Features: lag\_1 through lag\_7 (past 7 weeks), weekofyear (1--52), holiday\_count\\
\hspace*{1em}-- Target: weekly global active power (kWh)\\[2pt]
\textbf{Model Used:} XGBoost Regressor\\
\textbf{Model Performance:} MAE: 20.55, RMSE: 25.03, R2: 0.69\\[2pt]
\textbf{Prediction:} 175.44 kWh\\[2pt]
\textbf{Instance Features:}\\
\hspace*{1em}lag\_1: 163.10, lag\_2: 158.90, lag\_3: 171.20, ..., weekofyear: 42, holiday\_count: 1\\[2pt]
\textbf{SHAP Values:}\\
\hspace*{1em}lag\_1: +12.34, lag\_2: -5.21, lag\_3: +3.45, weekofyear: -8.12, holiday\_count: +2.89\\
\textbf{Base Value:} 152.00 kWh
\end{tcolorbox}

\subsection{Generated Explanation}

\begin{tcolorbox}[assistantprompt, title={Generated NLE (Zero-shot)}]
-- \textbf{Predicted consumption}: The model forecasts \textbf{175.44 kWh} for next week, about 15\% above the typical weekly average of 152 kWh.\\[2pt]
-- \textbf{Main driver}: Last week's consumption (163.10 kWh) is the strongest factor pushing the forecast up by about 12 kWh. When recent usage is high, the model expects that pattern to continue.\\[2pt]
-- \textbf{Seasonal adjustment}: Week 42 (mid-October) typically has lower consumption, which pulls the forecast down by about 8 kWh. Autumn weather may reduce heating/cooling needs.\\[2pt]
-- \textbf{Holiday effect}: One public holiday this week adds roughly 3 kWh to the forecast, as holidays often correlate with more time spent at home.\\[2pt]
-- \textbf{Model reliability}: With an R2 of 0.69, the model explains about 69\% of the variation in weekly consumption. The average error is around 20--25 kWh, so actual consumption could reasonably fall between 150--200 kWh.\\[2pt]
-- \textbf{Bottom line}: Expect moderately high consumption driven by recent usage patterns, partially offset by the season. Plan for some uncertainty given the model's typical error range.
\end{tcolorbox}

\section{Evaluation Rubrics}
\label{app:rubrics}

Each G-Eval dimension uses chain-of-thought evaluation steps followed by a 1--5 score.

\subsection{Accuracy Rubric}

\begin{tcolorbox}[rubricbox, title=Accuracy (1--5)]
You are evaluating the ACCURACY of a time-series forecast explanation.\\[4pt]
\textit{Context: Time series data with single-step ahead prediction of weekly global active power in kWh.}\\[4pt]
\textbf{Scoring:}\\
\hspace*{1em}5: All provided information is accurately represented (numerical values, feature names, relationships, metrics)\\
\hspace*{1em}4: Minor inaccuracies in non-critical details but core provided facts are correct\\
\hspace*{1em}3: Some factual errors in representing provided data but main message intact\\
\hspace*{1em}2: Multiple factual errors that misrepresent the provided information\\
\hspace*{1em}1: Major hallucinations or significant misrepresentation of provided data\\[4pt]
\textbf{Chain of Thought Steps:}\\
\hspace*{1em}1. Identify what information was actually provided in the context\\
\hspace*{1em}2. Check if mentioned numerical values match the provided data\\
\hspace*{1em}3. Verify that any referenced feature names and values are correctly stated\\
\hspace*{1em}4. Confirm that any mentioned model performance metrics are accurately reported\\
\hspace*{1em}5. Assess if relationships described are factually correct based on provided information\\
\hspace*{1em}6. Rate overall factual accuracy of representing the PROVIDED information only\\[4pt]
\textbf{Important:} Do not penalize explanations for missing information if none was provided in the context. Only evaluate accuracy of what information was actually given.
\end{tcolorbox}

\subsection{Lay User Relevancy Rubric}

\begin{tcolorbox}[rubricbox, title=Lay User Relevancy (1--5)]
You are a NON-TECHNICAL USER evaluating a forecast explanation.\\[4pt]
\textit{You have NO background in machine learning, statistics, or technical methods. You just want to understand why the system made this prediction and whether you should trust it.}\\[4pt]
\textbf{Scoring:}\\
\hspace*{1em}5: Extremely relevant and helpful for understanding and trusting the prediction\\
\hspace*{1em}4: Very relevant with useful information for my needs\\
\hspace*{1em}3: Somewhat relevant but could be more helpful\\
\hspace*{1em}2: Limited relevance to what I need to know\\
\hspace*{1em}1: Not relevant or confusing for a non-technical user\\[4pt]
\textbf{Chain of Thought Steps:}\\
\hspace*{1em}1. Does this explanation help me understand WHY the prediction was made?\\
\hspace*{1em}2. Can I determine if I should trust this prediction based on the explanation?\\
\hspace*{1em}3. Is the information presented in a way I can understand?\\
\hspace*{1em}4. Does this give me actionable insights for my decision-making?\\
\hspace*{1em}5. Rate overall relevancy for a lay user
\end{tcolorbox}

\subsection{Expert Relevancy Rubric}

\begin{tcolorbox}[rubricbox, title=Expert Relevancy (1--5)]
You are a MACHINE LEARNING EXPERT evaluating a forecast explanation.\\[4pt]
\textit{You have deep knowledge of machine learning models and time-series forecasting methods.}\\[4pt]
\textbf{Scoring:}\\
\hspace*{1em}5: Extremely relevant with appropriate technical depth for the information provided\\
\hspace*{1em}4: Very relevant with good technical information\\
\hspace*{1em}3: Somewhat relevant but missing some key technical aspects\\
\hspace*{1em}2: Limited technical relevance\\
\hspace*{1em}1: Not technically relevant or misleading\\[4pt]
\textbf{Chain of Thought Steps:}\\
\hspace*{1em}1. Does the explanation provide sufficient technical reasoning given the available information?\\
\hspace*{1em}2. Can I assess model behavior and prediction reliability from this explanation?\\
\hspace*{1em}3. Is the technical content accurate and well-presented?\\
\hspace*{1em}4. Does this meet expert needs given available information?\\
\hspace*{1em}5. Rate overall technical relevancy
\end{tcolorbox}

\subsection{Explanation Helpfulness Rubric}

\begin{tcolorbox}[rubricbox, title=Explanation Helpfulness (1--5)]
You are a NON-TECHNICAL USER evaluating how useful a forecast explanation is for judging prediction accuracy.\\[4pt]
\textit{You have NO background in machine learning or statistics. You just want to know if the explanation gives you enough information to decide how close the prediction is to the true value.}\\[4pt]
\textbf{Scoring:}\\
\hspace*{1em}5: Makes it very clear how accurate the prediction is\\
\hspace*{1em}4: Gives good guidance about prediction accuracy\\
\hspace*{1em}3: Provides some helpful information\\
\hspace*{1em}2: Gives limited guidance\\
\hspace*{1em}1: Does not help assess accuracy at all\\[4pt]
\textbf{Chain of Thought Steps:}\\
\hspace*{1em}1. Check clarity: Is the explanation understandable for a lay user (plain language, concrete references)?\\
\hspace*{1em}2. Check evidence: Does it include cues that inform accuracy (e.g., recent error rates, uncertainty ranges, comparison to history/seasonality, drivers affecting reliability)?\\
\hspace*{1em}3. Check actionability: Can a lay user form a reasonable judgment about closeness from the information given?\\
\hspace*{1em}4. Penalize jargon-only or vague statements that do not improve judgment\\
\hspace*{1em}5. Assign a score using the 1--5 scale above
\end{tcolorbox}

\section{Statistical Analysis Details}
\label{app:statistical}

This appendix provides complete statistical analysis for each research question. To ensure rigor and reproducibility, we apply a \textbf{standardized analysis protocol} across all RQs (except RQ4, which requires a specialized design due to SARIMAX constraints).

\paragraph{Evaluation Dimensions.} We evaluate NLE quality across four dimensions: \textbf{Accuracy} (faithfulness to XAI outputs), \textbf{Lay Relevancy} (accessibility for non-experts), \textbf{Expert Relevancy} (technical depth), and \textbf{Helpfulness} (overall usefulness). Each dimension uses a 1--5 scale; scores are averaged across both judges (GPT-4 and DeepSeek-R1) to mitigate self-preference bias.

\paragraph{Standardized Analysis Pipeline.} For each RQ, we apply the following sequence:

\begin{enumerate}[leftmargin=*]
    \item \textbf{One-way ANOVA} with omega-squared ($\omega^2$) effect sizes \citep{kroes2023demystifying}. We report $F$-statistics, raw $p$-values, FDR-corrected $p$-values \citep{benjamini1995controlling}, and effect size classifications.

    \item \textbf{Pairwise comparisons} using independent $t$-tests with Cohen's $d$ effect sizes \citep{cohen1988statistical} and Benjamini-Hochberg FDR correction applied within each dimension.

    \item \textbf{Welch ANOVA} as robustness check \citep{welch1951comparison}, which does not assume equal variances across groups -- important given our unbalanced designs (e.g., CoT strategies exclude DeepSeek-R1).

    \item \textbf{Games-Howell post-hoc tests} \citep{games1976pairwise} for significant dimensions, which are robust to unequal variances and sample sizes.

    \item \textbf{Factorial linear model} (Type II SS) controlling for other experimental factors, testing whether the main effect persists after accounting for confounds.

    \item \textbf{Interaction analysis} testing whether the main factor's effect varies across levels of other factors (e.g., does XAI effect differ by LLM?). We test all two-way interactions: Factor $\times$ LLM, Factor $\times$ Model, Factor $\times$ XAI, and Factor $\times$ Strategy (where applicable).

    \item \textbf{Simple effects analysis} for significant interactions, examining the main factor's effect within each level of the moderating factor.
\end{enumerate}

\paragraph{Effect Size Interpretation.} Following \citet{cohen1988statistical} and \citet{kroes2023demystifying}:
\begin{itemize}[leftmargin=*]
    \item $\omega^2$ (omnibus): $<.01$ negligible, $.01$--$.06$ small, $.06$--$.14$ medium, $>.14$ large
    \item Cohen's $d$ / Hedges' $g$ (pairwise): $<.2$ negligible, $.2$--$.5$ small, $.5$--$.8$ medium, $>.8$ large
\end{itemize}

\paragraph{Multiple Comparison Correction.} We apply Benjamini-Hochberg FDR correction \citep{benjamini1995controlling} to control false discovery rate at $\alpha=.05$. FDR correction is applied: (1) across dimensions within each ANOVA, and (2) within each dimension for pairwise comparisons.

\paragraph{RQ4 Exception.} RQ4 (SARIMAX vs.\ ML models) uses a specialized design because SARIMAX only supports XAI=``none.'' Fair comparisons filter to this condition ($N=264$), and we focus on testing the ``interpretability paradox'' hypothesis rather than following the full interaction protocol.

\paragraph{RQ5 Supplementary Analysis.} For RQ5 (prompting strategy), we supplement pointwise G-Eval with \textit{listwise ranking evaluation}. Both judges ranked all 8 strategies for 60 matched sets, validated via Friedman test and correlation with G-Eval scores.

\subsection{RQ1: XAI Effect}
\label{app:stat_rq1}

Analysis excludes SARIMAX (only supports XAI=``none''), yielding $N=594$ with balanced groups (198 per XAI condition).

\paragraph{ANOVA Results.} Table~\ref{tab:rq1_anova} shows the omnibus test for XAI effect across all four dimensions.

\begin{table}[H]
\centering
\small
\resizebox{\columnwidth}{!}{%
\begin{tabular}{lcccccc}
\toprule
\textbf{Dimension} & \textbf{F} & \textbf{p} & \textbf{p\textsubscript{FDR}} & \textbf{$\omega^2$} & \textbf{Effect} & \textbf{Sig.} \\
\midrule
Accuracy & 1.07 & .342 & .456 & .000 & negl. & \\
Lay Relevancy & 0.00 & .999 & .999 & .000 & negl. & \\
Expert Relevancy & 6.32 & .002 & .008 & .018 & small & ** \\
Helpfulness & 1.57 & .209 & .418 & .002 & negl. & \\
\bottomrule
\end{tabular}%
}
\caption{One-way ANOVA for XAI effect (RQ1). Only Expert Relevancy is significant.}
\label{tab:rq1_anova}
\end{table}

\paragraph{Pairwise Comparisons.} Table~\ref{tab:rq1_pairwise} shows pairwise $t$-tests with Cohen's $d$ effect sizes and FDR-corrected $p$-values for all dimensions.

\begin{table}[H]
\centering
\small
\resizebox{\columnwidth}{!}{%
\begin{tabular}{llccccc}
\toprule
\textbf{Dimension} & \textbf{Comparison} & \textbf{$\Delta$} & \textbf{$d$} & \textbf{Effect} & \textbf{p} & \textbf{p\textsubscript{FDR}} \\
\midrule
\multirow{3}{*}{Accuracy} & SHAP vs None & +0.04 & +0.08 & negl. & .445 & .479 \\
 & LIME vs None & +0.08 & +0.15 & negl. & .149 & .448 \\
 & SHAP vs LIME & $-$0.04 & $-$0.07 & negl. & .479 & .479 \\
\midrule
\multirow{3}{*}{Lay Rel.} & SHAP vs None & +0.00 & +0.01 & negl. & .960 & .996 \\
 & LIME vs None & +0.00 & +0.00 & negl. & .996 & .996 \\
 & SHAP vs LIME & +0.00 & +0.00 & negl. & .965 & .996 \\
\midrule
\multirow{3}{*}{Expert Rel.} & SHAP vs None & +0.19 & +0.33 & small & .001 & .003* \\
 & LIME vs None & +0.16 & +0.28 & small & .006 & .008* \\
 & SHAP vs LIME & +0.03 & +0.06 & negl. & .558 & .558 \\
\midrule
\multirow{3}{*}{Helpfulness} & SHAP vs None & $-$0.08 & $-$0.16 & negl. & .108 & .221 \\
 & LIME vs None & $-$0.07 & $-$0.15 & negl. & .147 & .221 \\
 & SHAP vs LIME & $-$0.01 & $-$0.02 & negl. & .822 & .822 \\
\bottomrule
\end{tabular}%
}
\caption{Pairwise comparisons for RQ1 (XAI effect). FDR correction applied within each dimension.}
\label{tab:rq1_pairwise}
\end{table}

\paragraph{Robustness Check: Welch ANOVA.} Table~\ref{tab:rq1_welch} shows Welch's ANOVA, which is robust to heterogeneous variances. Results confirm standard ANOVA findings.

\begin{table}[H]
\centering
\small
\resizebox{\columnwidth}{!}{%
\begin{tabular}{lccc}
\toprule
\textbf{Dimension} & \textbf{F} & \textbf{p} & \textbf{Sig.} \\
\midrule
Accuracy & 1.04 & .354 & \\
Lay Relevancy & 0.00 & .999 & \\
Expert Relevancy & 6.45 & .002 & ** \\
Helpfulness & 1.60 & .203 & \\
\bottomrule
\end{tabular}%
}
\caption{Welch ANOVA for RQ1 (robust to unequal variances).}
\label{tab:rq1_welch}
\end{table}

\paragraph{Post-hoc Tests.} Table~\ref{tab:rq1_gameshowell} shows Games-Howell post-hoc results for Expert Relevancy (the only significant dimension).

\begin{table}[H]
\centering
\small
\resizebox{\columnwidth}{!}{%
\begin{tabular}{llcccccc}
\toprule
\textbf{A} & \textbf{B} & \textbf{Mean(A)} & \textbf{Mean(B)} & \textbf{Diff} & \textbf{SE} & \textbf{p} & \textbf{Hedges' $g$} \\
\midrule
LIME & None & 3.94 & 3.78 & +0.16 & 0.057 & .016 & +0.28 \\
LIME & SHAP & 3.94 & 3.97 & $-$0.03 & 0.059 & .827 & $-$0.06 \\
None & SHAP & 3.78 & 3.97 & $-$0.19 & 0.058 & .003 & $-$0.33 \\
\bottomrule
\end{tabular}%
}
\caption{Games-Howell for Expert Relevancy (RQ1). Robust to unequal variances.}
\label{tab:rq1_gameshowell}
\end{table}

\paragraph{Factorial Model.} Table~\ref{tab:rq1_factorial} shows XAI effect when controlling for LLM and Strategy (Type II SS).

\begin{table}[H]
\centering
\small
\resizebox{\columnwidth}{!}{%
\begin{tabular}{lccc}
\toprule
\textbf{Dimension} & \textbf{XAI F} & \textbf{XAI p} & \textbf{Sig.} \\
\midrule
Accuracy & 1.57 & .209 & \\
Lay Relevancy & 0.00 & .998 & \\
Expert Relevancy & 12.93 & $<$.001 & *** \\
Helpfulness & 2.98 & .051 & \\
\bottomrule
\end{tabular}%
}
\caption{Factorial model: XAI effect controlling for LLM and Strategy (RQ1).}
\label{tab:rq1_factorial}
\end{table}

\paragraph{Interaction Analysis.} Tables~\ref{tab:rq1_interaction_llm}--\ref{tab:rq1_interaction_strategy} show all two-way interactions for the XAI factor.

\begin{table}[H]
\centering
\small
\resizebox{\columnwidth}{!}{%
\begin{tabular}{lcccc}
\toprule
\textbf{Dimension} & \textbf{F} & \textbf{p} & \textbf{$\omega^2$} & \textbf{Sig.} \\
\midrule
Accuracy & 2.25 & .062 & .008 & \\
Lay Relevancy & 3.05 & .017 & .014 & * \\
Expert Relevancy & 2.68 & .031 & .011 & * \\
Helpfulness & 2.00 & .093 & .007 & \\
\bottomrule
\end{tabular}%
}
\caption{XAI $\times$ LLM interaction tests (RQ1).}
\label{tab:rq1_interaction_llm}
\end{table}

\begin{table}[H]
\centering
\small
\resizebox{\columnwidth}{!}{%
\begin{tabular}{lcccc}
\toprule
\textbf{Dimension} & \textbf{F} & \textbf{p} & \textbf{$\omega^2$} & \textbf{Sig.} \\
\midrule
Accuracy & 0.39 & .818 & .000 & \\
Lay Relevancy & 0.97 & .424 & .000 & \\
Expert Relevancy & 0.12 & .974 & .000 & \\
Helpfulness & 0.91 & .456 & .000 & \\
\bottomrule
\end{tabular}%
}
\caption{XAI $\times$ Model interaction tests (RQ1). No significant interactions.}
\label{tab:rq1_interaction_model}
\end{table}

\begin{table}[H]
\centering
\small
\resizebox{\columnwidth}{!}{%
\begin{tabular}{lcccc}
\toprule
\textbf{Dimension} & \textbf{F} & \textbf{p} & \textbf{$\omega^2$} & \textbf{Sig.} \\
\midrule
Accuracy & 1.32 & .189 & .008 & \\
Lay Relevancy & 0.67 & .800 & .000 & \\
Expert Relevancy & 1.05 & .403 & .001 & \\
Helpfulness & 0.40 & .976 & .000 & \\
\bottomrule
\end{tabular}%
}
\caption{XAI $\times$ Strategy interaction tests (RQ1). No significant interactions.}
\label{tab:rq1_interaction_strategy}
\end{table}

\paragraph{Simple Effects.} Table~\ref{tab:rq1_simple} shows XAI effect within each LLM for dimensions with significant interactions.

\begin{table}[H]
\centering
\small
\resizebox{\columnwidth}{!}{%
\begin{tabular}{llccccc}
\toprule
\textbf{Dimension} & \textbf{LLM} & \textbf{F} & \textbf{p} & \textbf{SHAP} & \textbf{LIME} & \textbf{None} \\
\midrule
\multirow{3}{*}{Expert Rel.} & GPT-4o & 16.01 & $<$.001* & 4.13 & 4.01 & 3.79 \\
 & Llama-3 & 1.36 & .258 & 3.40 & 3.49 & 3.35 \\
 & DeepSeek & 4.64 & .011* & 4.51 & 4.44 & 4.33 \\
\midrule
\multirow{3}{*}{Lay Rel.} & GPT-4o & 0.11 & .893 & 4.23 & 4.23 & 4.26 \\
 & Llama-3 & 1.40 & .248 & 4.01 & 4.03 & 4.15 \\
 & DeepSeek & 7.05 & .001* & 4.84 & 4.81 & 4.61 \\
\bottomrule
\end{tabular}%
}
\caption{Simple effects: XAI effect within each LLM (RQ1).}
\label{tab:rq1_simple}
\end{table}

\subsection{RQ2: LLM Effect}
\label{app:stat_rq2}

Analysis uses full dataset ($N=660$). Note unbalanced design: DeepSeek-R1 has 180 samples (CoT strategies excluded as native reasoning model), while GPT-4o and Llama-3 have 240 each.

\paragraph{ANOVA Results.} Table~\ref{tab:rq2_anova} shows the omnibus test for LLM effect. All four dimensions show highly significant effects with large effect sizes.

\begin{table}[H]
\centering
\small
\resizebox{\columnwidth}{!}{%
\begin{tabular}{lcccccc}
\toprule
\textbf{Dimension} & \textbf{F} & \textbf{p} & \textbf{p\textsubscript{FDR}} & \textbf{$\omega^2$} & \textbf{Effect} & \textbf{Sig.} \\
\midrule
Accuracy & 81.32 & $<$.001 & $<$.001 & .196 & large & *** \\
Lay Relevancy & 107.85 & $<$.001 & $<$.001 & .245 & large & *** \\
Expert Relevancy & 332.15 & $<$.001 & $<$.001 & .501 & large & *** \\
Helpfulness & 206.35 & $<$.001 & $<$.001 & .384 & large & *** \\
\bottomrule
\end{tabular}%
}
\caption{One-way ANOVA for LLM effect (RQ2). Large effects on all dimensions.}
\label{tab:rq2_anova}
\end{table}

\paragraph{Pairwise Comparisons.} Table~\ref{tab:rq2_pairwise} shows pairwise $t$-tests with Cohen's $d$ and FDR-corrected $p$-values.

\begin{table}[H]
\centering
\small
\resizebox{\columnwidth}{!}{%
\begin{tabular}{llccccc}
\toprule
\textbf{Dimension} & \textbf{Comparison} & \textbf{$\Delta$} & \textbf{$d$} & \textbf{Effect} & \textbf{p} & \textbf{p\textsubscript{FDR}} \\
\midrule
\multirow{3}{*}{Accuracy} & DeepSeek vs GPT & $-$0.02 & $-$0.04 & negl. & .680 & .680 \\
 & DeepSeek vs Llama & +0.53 & +0.91 & large & $<$.001 & $<$.001* \\
 & GPT vs Llama & +0.54 & +0.95 & large & $<$.001 & $<$.001* \\
\midrule
\multirow{3}{*}{Lay Rel.} & DeepSeek vs GPT & +0.51 & +1.17 & large & $<$.001 & $<$.001* \\
 & DeepSeek vs Llama & +0.67 & +1.40 & large & $<$.001 & $<$.001* \\
 & GPT vs Llama & +0.16 & +0.31 & small & .001 & .001* \\
\midrule
\multirow{3}{*}{Expert Rel.} & DeepSeek vs GPT & +0.50 & +1.33 & large & $<$.001 & $<$.001* \\
 & DeepSeek vs Llama & +1.06 & +2.50 & large & $<$.001 & $<$.001* \\
 & GPT vs Llama & +0.56 & +1.24 & large & $<$.001 & $<$.001* \\
\midrule
\multirow{3}{*}{Helpfulness} & DeepSeek vs GPT & +0.50 & +1.55 & large & $<$.001 & $<$.001* \\
 & DeepSeek vs Llama & +0.79 & +1.91 & large & $<$.001 & $<$.001* \\
 & GPT vs Llama & +0.28 & +0.66 & medium & $<$.001 & $<$.001* \\
\bottomrule
\end{tabular}%
}
\caption{Pairwise comparisons for RQ2 (LLM effect). FDR correction applied within each dimension.}
\label{tab:rq2_pairwise}
\end{table}

\paragraph{Robustness Check: Welch ANOVA.} Table~\ref{tab:rq2_welch} shows Welch's ANOVA (robust to unequal variances and sample sizes). Results confirm standard ANOVA.

\begin{table}[H]
\centering
\small
\resizebox{\columnwidth}{!}{%
\begin{tabular}{lccc}
\toprule
\textbf{Dimension} & \textbf{F} & \textbf{p} & \textbf{Sig.} \\
\midrule
Accuracy & 59.81 & $<$.001 & *** \\
Lay Relevancy & 128.21 & $<$.001 & *** \\
Expert Relevancy & 363.22 & $<$.001 & *** \\
Helpfulness & 261.40 & $<$.001 & *** \\
\bottomrule
\end{tabular}%
}
\caption{Welch ANOVA for RQ2 (robust to unequal variances/samples).}
\label{tab:rq2_welch}
\end{table}

\paragraph{Post-hoc Tests.} Games-Howell for Expert Relevancy (strongest effect): DeepSeek vs GPT ($\Delta=0.50$, $p<.001$, Hedges' $g=1.33$), DeepSeek vs Llama ($\Delta=1.06$, $p<.001$, $g=2.50$), GPT vs Llama ($\Delta=0.56$, $p<.001$, $g=1.24$).

\paragraph{Factorial Model.} Table~\ref{tab:rq2_factorial} shows LLM effect when controlling for XAI and Strategy (Type II SS). Effects remain highly significant.

\begin{table}[H]
\centering
\small
\resizebox{\columnwidth}{!}{%
\begin{tabular}{lccc}
\toprule
\textbf{Dimension} & \textbf{LLM F} & \textbf{LLM p} & \textbf{Sig.} \\
\midrule
Accuracy & 95.70 & $<$.001 & *** \\
Lay Relevancy & 112.58 & $<$.001 & *** \\
Expert Relevancy & 334.24 & $<$.001 & *** \\
Helpfulness & 220.51 & $<$.001 & *** \\
\bottomrule
\end{tabular}%
}
\caption{Factorial model: LLM effect controlling for XAI and Strategy (RQ2).}
\label{tab:rq2_factorial}
\end{table}

\paragraph{Interaction Analysis.} Tables~\ref{tab:rq2_interaction_strategy}--\ref{tab:rq2_interaction_model} show all two-way interactions for the LLM factor.

\begin{table}[H]
\centering
\small
\resizebox{\columnwidth}{!}{%
\begin{tabular}{lcccc}
\toprule
\textbf{Dimension} & \textbf{F} & \textbf{p} & \textbf{$\omega^2$} & \textbf{Sig.} \\
\midrule
Accuracy & 18.79 & $<$.001 & .276 & *** \\
Lay Relevancy & 19.81 & $<$.001 & .287 & *** \\
Expert Relevancy & 23.23 & $<$.001 & .323 & *** \\
Helpfulness & 33.18 & $<$.001 & .408 & *** \\
\bottomrule
\end{tabular}%
}
\caption{LLM $\times$ Strategy interaction tests (RQ2). All significant.}
\label{tab:rq2_interaction_strategy}
\end{table}

\begin{table}[H]
\centering
\small
\resizebox{\columnwidth}{!}{%
\begin{tabular}{lcccc}
\toprule
\textbf{Dimension} & \textbf{F} & \textbf{p} & \textbf{$\omega^2$} & \textbf{Sig.} \\
\midrule
Accuracy & 1.60 & .172 & .004 & \\
Lay Relevancy & 3.17 & .014 & .013 & * \\
Expert Relevancy & 3.63 & .006 & .016 & ** \\
Helpfulness & 2.22 & .066 & .007 & \\
\bottomrule
\end{tabular}%
}
\caption{LLM $\times$ XAI interaction tests (RQ2). Small effects on 2 dimensions.}
\label{tab:rq2_interaction_xai}
\end{table}

\begin{table}[H]
\centering
\small
\resizebox{\columnwidth}{!}{%
\begin{tabular}{lcccc}
\toprule
\textbf{Dimension} & \textbf{F} & \textbf{p} & \textbf{$\omega^2$} & \textbf{Sig.} \\
\midrule
Accuracy & 1.08 & .374 & .001 & \\
Lay Relevancy & 1.06 & .387 & .001 & \\
Expert Relevancy & 3.68 & .001 & .024 & ** \\
Helpfulness & 0.38 & .891 & .000 & \\
\bottomrule
\end{tabular}%
}
\caption{LLM $\times$ Model interaction tests (RQ2). Expert Relevancy shows significant interaction.}
\label{tab:rq2_interaction_model}
\end{table}

\paragraph{Simple Effects.} LLM effect within each Strategy for Helpfulness (representative dimension): All strategies show significant LLM effects ($p<.05$). DeepSeek-R1 consistently achieves highest scores when available (excluded from CoT strategies). Within CoT strategies (GPT vs Llama only), GPT-4o outperforms Llama-3.

\subsection{RQ3: ML Model Effect}
\label{app:stat_rq3}

Analysis uses ML models only ($N=594$): XGBoost ($N=198$), Random Forest ($N=198$), MLP ($N=198$). SARIMAX is analyzed separately in RQ4.

\paragraph{ANOVA Results.} Table~\ref{tab:rq3_anova} shows the omnibus test for Model effect across three ML models.

\begin{table}[H]
\centering
\small
\resizebox{\columnwidth}{!}{%
\begin{tabular}{lcccccc}
\toprule
\textbf{Dimension} & \textbf{F} & \textbf{p} & \textbf{p\textsubscript{FDR}} & \textbf{$\omega^2$} & \textbf{Effect} & \textbf{Sig.} \\
\midrule
Accuracy & 0.60 & .550 & .550 & .000 & negl. & \\
Lay Relevancy & 40.12 & $<$.001 & $<$.001 & .117 & medium & *** \\
Expert Relevancy & 5.41 & .005 & .006 & .015 & small & ** \\
Helpfulness & 5.70 & .004 & .005 & .016 & small & ** \\
\bottomrule
\end{tabular}%
}
\caption{One-way ANOVA for ML Model effect (RQ3). $N=594$, $df=(2,591)$.}
\label{tab:rq3_anova}
\end{table}

\paragraph{Pairwise Comparisons.} Table~\ref{tab:rq3_pairwise} shows pairwise $t$-tests with Cohen's $d$ effect sizes and FDR-corrected $p$-values for all dimensions.

\begin{table}[H]
\centering
\small
\resizebox{\columnwidth}{!}{%
\begin{tabular}{llccccc}
\toprule
\textbf{Dimension} & \textbf{Comparison} & \textbf{$\Delta$} & \textbf{$d$} & \textbf{Effect} & \textbf{p} & \textbf{p\textsubscript{FDR}} \\
\midrule
\multirow{3}{*}{Lay Rel.} & XGB vs RF & +0.25 & +0.48 & small & $<$.001 & $<$.001* \\
 & XGB vs MLP & +0.41 & +0.82 & large & $<$.001 & $<$.001* \\
 & RF vs MLP & +0.16 & +0.30 & small & .003 & .004* \\
\midrule
\multirow{3}{*}{Expert Rel.} & XGB vs RF & +0.18 & +0.33 & small & .001 & .003* \\
 & XGB vs MLP & +0.14 & +0.24 & small & .024 & .036* \\
 & RF vs MLP & $-$0.04 & $-$0.07 & negl. & .452 & .452 \\
\midrule
\multirow{3}{*}{Helpfulness} & XGB vs RF & +0.16 & +0.32 & small & .001 & .004* \\
 & XGB vs MLP & +0.14 & +0.29 & small & .007 & .014* \\
 & RF vs MLP & $-$0.02 & $-$0.04 & negl. & .681 & .681 \\
\bottomrule
\end{tabular}%
}
\caption{Pairwise comparisons for RQ3 (Model effect). FDR correction applied within each dimension.}
\label{tab:rq3_pairwise}
\end{table}

\paragraph{Robustness Check: Welch ANOVA.} Table~\ref{tab:rq3_welch} confirms results using Welch's ANOVA (robust to variance heterogeneity).

\begin{table}[H]
\centering
\small
\resizebox{\columnwidth}{!}{%
\begin{tabular}{lccc}
\toprule
\textbf{Dimension} & \textbf{F} & \textbf{p} & \textbf{Sig.} \\
\midrule
Accuracy & 0.59 & .554 & \\
Lay Relevancy & 40.28 & $<$.001 & *** \\
Expert Relevancy & 5.28 & .005 & ** \\
Helpfulness & 5.68 & .004 & ** \\
\bottomrule
\end{tabular}%
}
\caption{Welch ANOVA for RQ3 (ML models only).}
\label{tab:rq3_welch}
\end{table}

\paragraph{Games-Howell Post-hoc.} Robust to unequal variances. Significant pairs: Lay Relevancy (3): XGB$>$RF ($g=0.48$), XGB$>$MLP ($g=0.82$), RF$>$MLP ($g=0.30$). Expert Relevancy (1): XGB$>$RF ($g=0.33$). Helpfulness (2): XGB$>$MLP ($g=0.29$), XGB$>$RF ($g=0.32$). Confirms Tukey HSD findings.

\paragraph{Factorial Model.} Table~\ref{tab:rq3_factorial} shows Model effect controlling for LLM, XAI, and Strategy (Type II SS).

\begin{table}[H]
\centering
\small
\resizebox{\columnwidth}{!}{%
\begin{tabular}{lcccc}
\toprule
\textbf{Dimension} & \textbf{F} & \textbf{p} & \textbf{$\eta^2_p$} & \textbf{Sig.} \\
\midrule
Accuracy & 1.02 & .360 & .004 & \\
Lay Relevancy & 51.94 & $<$.001 & .152 & *** \\
Expert Relevancy & 11.65 & $<$.001 & .039 & *** \\
Helpfulness & 11.89 & $<$.001 & .039 & *** \\
\bottomrule
\end{tabular}%
}
\caption{Factorial model: Model effect controlling for confounds (RQ3).}
\label{tab:rq3_factorial}
\end{table}

\paragraph{Interaction Tests.} Tables~\ref{tab:rq3_interaction_llm}--\ref{tab:rq3_interaction_strategy} show all two-way interactions for the Model factor.

\begin{table}[H]
\centering
\small
\resizebox{\columnwidth}{!}{%
\begin{tabular}{lcccc}
\toprule
\textbf{Dimension} & \textbf{F} & \textbf{p} & \textbf{$\eta^2_p$} & \textbf{Sig.} \\
\midrule
Accuracy & 0.67 & .615 & .005 & \\
Lay Relevancy & 1.12 & .347 & .008 & \\
Expert Relevancy & 1.66 & .157 & .011 & \\
Helpfulness & 0.16 & .958 & .001 & \\
\bottomrule
\end{tabular}%
}
\caption{Model $\times$ LLM interaction tests (RQ3). No significant interactions.}
\label{tab:rq3_interaction_llm}
\end{table}

\begin{table}[H]
\centering
\small
\resizebox{\columnwidth}{!}{%
\begin{tabular}{lcccc}
\toprule
\textbf{Dimension} & \textbf{F} & \textbf{p} & \textbf{$\eta^2_p$} & \textbf{Sig.} \\
\midrule
Accuracy & 0.39 & .818 & .003 & \\
Lay Relevancy & 0.97 & .424 & .007 & \\
Expert Relevancy & 0.12 & .974 & .001 & \\
Helpfulness & 0.91 & .456 & .006 & \\
\bottomrule
\end{tabular}%
}
\caption{Model $\times$ XAI interaction tests (RQ3). No significant interactions.}
\label{tab:rq3_interaction_xai}
\end{table}

\begin{table}[H]
\centering
\small
\resizebox{\columnwidth}{!}{%
\begin{tabular}{lcccc}
\toprule
\textbf{Dimension} & \textbf{F} & \textbf{p} & \textbf{$\eta^2_p$} & \textbf{Sig.} \\
\midrule
Accuracy & 0.68 & .794 & .016 & \\
Lay Relevancy & 0.77 & .704 & .019 & \\
Expert Relevancy & 0.44 & .960 & .011 & \\
Helpfulness & 0.50 & .936 & .012 & \\
\bottomrule
\end{tabular}%
}
\caption{Model $\times$ Strategy interaction tests (RQ3). No significant interactions.}
\label{tab:rq3_interaction_strategy}
\end{table}

No significant interactions (all $p>.05$), indicating Model effects are consistent across LLMs and Strategies.

\paragraph{Simple Effects.} Model effect within each LLM for Lay Relevancy (representative dimension): GPT ($F=29.0$, $p<.001$, $\omega^2=.21$), DeepSeek ($F=15.0$, $p<.001$, $\omega^2=.15$), Llama ($F=8.7$, $p<.001$, $\omega^2=.07$). All LLMs show significant Model effects ($p<.05$) for Lay Relevancy and Expert Relevancy. For Helpfulness, GPT and DeepSeek show significant effects.

\paragraph{Model R$^2$ vs NLE Quality.} Spearman $\rho=1.00$ between model R$^2$ and overall NLE score ($n=3$, descriptive only). Higher-performing models yield better explanations.

\subsection{RQ4: SARIMAX (Classical Model)}
\label{app:stat_rq4}

Analysis compares SARIMAX (classical time series, $N=66$) against ML models. SARIMAX only supports XAI=``none'' in our design, so fair comparisons filter to XAI=``none'' only ($N=264$).

\paragraph{ANOVA at XAI=``none''.} Table~\ref{tab:rq4_anova} shows Model effect when all models have XAI=``none''.

\begin{table}[H]
\centering
\small
\resizebox{\columnwidth}{!}{%
\begin{tabular}{lcccccc}
\toprule
\textbf{Dimension} & \textbf{F} & \textbf{p} & \textbf{p\textsubscript{FDR}} & \textbf{$\omega^2$} & \textbf{Effect} & \textbf{Sig.} \\
\midrule
Accuracy & 0.11 & .952 & .952 & .000 & negl. & \\
Lay Relevancy & 19.24 & $<$.001 & $<$.001 & .172 & large & *** \\
Expert Relevancy & 1.40 & .244 & .305 & .004 & negl. & \\
Helpfulness & 8.89 & $<$.001 & $<$.001 & .082 & medium & *** \\
\bottomrule
\end{tabular}%
}
\caption{ANOVA for Model effect at XAI=``none'' (RQ4). $N=264$, $df=(3,260)$.}
\label{tab:rq4_anova}
\end{table}

\paragraph{Post-hoc: SARIMAX vs ML Models.} Table~\ref{tab:rq4_posthoc} compares SARIMAX to each ML model at XAI=``none''.

\begin{table}[H]
\centering
\small
\resizebox{\columnwidth}{!}{%
\begin{tabular}{llcccc}
\toprule
\textbf{Dimension} & \textbf{vs} & \textbf{$\Delta$} & \textbf{$d$} & \textbf{p} & \textbf{Sig.} \\
\midrule
Lay Relevancy & XGB & $-$0.57 & $-$1.16 & $<$.001 & *** \\
Lay Relevancy & RF & $-$0.23 & $-$0.44 & .012 & * \\
Lay Relevancy & MLP & $-$0.04 & $-$0.07 & .674 & \\
Helpfulness & XGB & $-$0.42 & $-$0.98 & $<$.001 & *** \\
Helpfulness & RF & $-$0.17 & $-$0.37 & .037 & * \\
Helpfulness & MLP & $-$0.19 & $-$0.41 & .020 & * \\
\bottomrule
\end{tabular}%
}
\caption{SARIMAX vs each ML model at XAI=``none'' (N=66 each). FDR-corrected.}
\label{tab:rq4_posthoc}
\end{table}

\paragraph{Testing the R$^2$ Hypothesis.} If prediction quality (R$^2$) drove NLE quality, SARIMAX (R$^2$=0.55) should outperform RF (R$^2$=0.37) and MLP (R$^2$=0.21). Table~\ref{tab:rq4_r2test} tests this hypothesis.

\begin{table}[H]
\centering
\small
\resizebox{\columnwidth}{!}{%
\begin{tabular}{llcccc}
\toprule
\textbf{Comparison} & \textbf{Dimension} & \textbf{$\Delta$R$^2$} & \textbf{$\Delta$NLE} & \textbf{$d$} & \textbf{p} \\
\midrule
SARIMAX vs MLP & Helpfulness & +0.34 & $-$0.19 & $-$0.41 & .020* \\
SARIMAX vs RF & Lay Relevancy & +0.18 & $-$0.23 & $-$0.44 & .012* \\
SARIMAX vs RF & Helpfulness & +0.18 & $-$0.17 & $-$0.37 & .037* \\
\bottomrule
\end{tabular}%
}
\caption{SARIMAX vs lower-R$^2$ models. $\Delta$R$^2$ = SARIMAX minus comparison model. Positive $\Delta$R$^2$ means SARIMAX has better predictions; negative $d$ means SARIMAX has worse NLEs.}
\label{tab:rq4_r2test}
\end{table}

Despite having \emph{higher} R$^2$, SARIMAX yields significantly \emph{worse} NLEs on multiple dimensions. This falsifies the R$^2$ hypothesis: prediction quality does not drive NLE quality.

\paragraph{SARIMAX vs Pooled ML.} Comparing SARIMAX ($N=66$) against all ML models ($N=594$, all XAI conditions): significant deficits on Lay Relevancy ($\Delta=-0.28$, $d=-0.51$, $p_{\text{FDR}}<.001$) and Helpfulness ($\Delta=-0.21$, $d=-0.42$, $p_{\text{FDR}}=.003$). Expert Relevancy shows marginal deficit ($d=-0.24$, $p=.07$). Accuracy unaffected.

\paragraph{LLM Effect Within SARIMAX.} Strong LLM effects: Expert Relevancy ($F=71.4$, $\omega^2=.68$), Helpfulness ($F=18.0$, $\omega^2=.34$), Accuracy ($F=15.6$, $\omega^2=.31$), Lay Relevancy ($F=7.5$, $\omega^2=.17$). DeepSeek-R1 achieves highest overall score (4.09), followed by GPT-4o (3.72), then Llama-3 (3.41).

\paragraph{Strategy Effect Within SARIMAX.} Modest Strategy effects on Lay Relevancy ($F=3.4$, $p=.005$, $\omega^2=.20$) and Accuracy ($F=2.5$, $p=.028$, $\omega^2=.13$). Top strategies: self-consistency (3.88), reflexion (3.86), role-based (3.76).

\paragraph{The Interpretability Paradox.} SARIMAX has inherently interpretable coefficients (AR, MA, seasonal), yet yields \emph{worse} NLEs than black-box ML models -- even models with \emph{lower} prediction quality. If R$^2$ were the driver, expected ranking would be: XGBoost (0.69) $>$ SARIMAX (0.55) $>$ RF (0.37) $>$ MLP (0.21). Actual NLE ranking: XGBoost $>$ RF $>$ MLP $>$ SARIMAX. This reveals that \textbf{model type matters more than prediction quality}: LLMs struggle to translate statistical coefficients (AR, MA terms) into accessible narratives, regardless of how accurate the predictions are.

\subsection{RQ5: Prompting Strategy Effect}
\label{app:stat_rq5}

Analysis uses full dataset ($N=660$). Strategies have unbalanced sample sizes due to design constraints: CoT strategies excluded DeepSeek-R1 (native reasoning model), yielding $N=60$ for CoT-zero and CoT-few vs.\ $N=90$ for other strategies.

\paragraph{ANOVA Results.} Table~\ref{tab:rq5_anova} shows the omnibus test for Strategy effect across all four dimensions.

\begin{table}[H]
\centering
\small
\resizebox{\columnwidth}{!}{%
\begin{tabular}{lcccccc}
\toprule
\textbf{Dimension} & \textbf{F} & \textbf{p} & \textbf{p\textsubscript{FDR}} & \textbf{$\omega^2$} & \textbf{Effect} & \textbf{Sig.} \\
\midrule
Accuracy & 16.75 & $<$.001 & $<$.001 & .143 & large & *** \\
Lay Relevancy & 14.28 & $<$.001 & $<$.001 & .123 & medium & *** \\
Expert Relevancy & 4.80 & $<$.001 & $<$.001 & .039 & small & *** \\
Helpfulness & 8.86 & $<$.001 & $<$.001 & .077 & medium & *** \\
\bottomrule
\end{tabular}%
}
\caption{One-way ANOVA for Strategy effect (RQ5). $N=660$, $df=(7,652)$.}
\label{tab:rq5_anova}
\end{table}

\paragraph{Robustness Check: Welch ANOVA.} Table~\ref{tab:rq5_welch} shows Welch's ANOVA, which is robust to heterogeneous variances and unequal sample sizes. Results confirm standard ANOVA findings.

\begin{table}[H]
\centering
\small
\resizebox{\columnwidth}{!}{%
\begin{tabular}{lccc}
\toprule
\textbf{Dimension} & \textbf{F} & \textbf{p} & \textbf{Sig.} \\
\midrule
Accuracy & 14.42 & $<$.001 & *** \\
Lay Relevancy & 16.47 & $<$.001 & *** \\
Expert Relevancy & 5.46 & $<$.001 & *** \\
Helpfulness & 9.79 & $<$.001 & *** \\
\bottomrule
\end{tabular}%
}
\caption{Welch ANOVA for RQ5 (robust to unequal variances/samples).}
\label{tab:rq5_welch}
\end{table}

\paragraph{Games-Howell Post-hoc.} Table~\ref{tab:rq5_gameshowell} shows significant pairwise comparisons (robust to unequal variances). Only pairs with $p_{\text{adj}}<.05$ shown.

\begin{table}[H]
\centering
\small
\resizebox{\columnwidth}{!}{%
\begin{tabular}{llcccc}
\toprule
\textbf{Dimension} & \textbf{Comparison} & \textbf{$\Delta$} & \textbf{Hedges' $g$} & \textbf{p\textsubscript{adj}} & \textbf{Sig.} \\
\midrule
Accuracy & Self-cons. vs CoT-few & +0.59 & +1.10 & $<$.001 & *** \\
Accuracy & Reflexion vs CoT-few & +0.61 & +1.14 & $<$.001 & *** \\
Accuracy & Zero-shot vs CoT-few & +0.57 & +1.10 & $<$.001 & *** \\
Accuracy & CoT-zero vs CoT-few & +0.58 & +1.04 & $<$.001 & *** \\
Accuracy & Self-cons. vs Meta & +0.50 & +0.93 & $<$.001 & *** \\
Lay Rel. & Self-cons. vs Meta & +0.60 & +1.23 & $<$.001 & *** \\
Lay Rel. & Self-cons. vs CoT-few & +0.55 & +1.02 & $<$.001 & *** \\
Lay Rel. & Role-based vs Meta & +0.51 & +1.01 & $<$.001 & *** \\
Lay Rel. & Reflexion vs Meta & +0.42 & +0.83 & .001 & ** \\
Expert Rel. & Self-cons. vs CoT-few & +0.51 & +0.85 & .003 & ** \\
Helpfulness & Self-cons. vs Meta & +0.37 & +0.80 & .005 & ** \\
Helpfulness & Few-shot vs Meta & +0.47 & +0.97 & $<$.001 & *** \\
\bottomrule
\end{tabular}%
}
\caption{Games-Howell significant pairs for Strategy (RQ5). FDR-corrected.}
\label{tab:rq5_gameshowell}
\end{table}

\paragraph{Factorial Model.} Table~\ref{tab:rq5_factorial} shows Strategy effect when controlling for LLM and XAI (Type II SS). Effects remain highly significant.

\begin{table}[H]
\centering
\small
\resizebox{\columnwidth}{!}{%
\begin{tabular}{lccc}
\toprule
\textbf{Dimension} & \textbf{Strategy F} & \textbf{p} & \textbf{Sig.} \\
\midrule
Accuracy & 20.80 & $<$.001 & *** \\
Lay Relevancy & 15.66 & $<$.001 & *** \\
Expert Relevancy & 3.55 & .001 & ** \\
Helpfulness & 12.06 & $<$.001 & *** \\
\bottomrule
\end{tabular}%
}
\caption{Factorial model: Strategy effect controlling for LLM and XAI (RQ5).}
\label{tab:rq5_factorial}
\end{table}

\paragraph{Two-way Interactions.} Tables~\ref{tab:rq5_int_llm}--\ref{tab:rq5_int_model} test all two-way interactions with Strategy.

\begin{table}[H]
\centering
\small
\resizebox{\columnwidth}{!}{%
\begin{tabular}{lcccc}
\toprule
\textbf{Dimension} & \textbf{F} & \textbf{p} & \textbf{$\omega^2$} & \textbf{Sig.} \\
\midrule
Accuracy & 18.79 & $<$.001 & .276 & *** \\
Lay Relevancy & 19.81 & $<$.001 & .287 & *** \\
Expert Relevancy & 23.23 & $<$.001 & .323 & *** \\
Helpfulness & 33.18 & $<$.001 & .408 & *** \\
\bottomrule
\end{tabular}%
}
\caption{Strategy $\times$ LLM interaction (RQ5). All dimensions show significant large effects.}
\label{tab:rq5_int_llm}
\end{table}

\begin{table}[H]
\centering
\small
\resizebox{\columnwidth}{!}{%
\begin{tabular}{lcccc}
\toprule
\textbf{Dimension} & \textbf{F} & \textbf{p} & \textbf{$\omega^2$} & \textbf{Sig.} \\
\midrule
Accuracy & 1.25 & .231 & .005 & -- \\
Lay Relevancy & 0.79 & .680 & .000 & -- \\
Expert Relevancy & 0.93 & .522 & .000 & -- \\
Helpfulness & 0.35 & .987 & .000 & -- \\
\bottomrule
\end{tabular}%
}
\caption{Strategy $\times$ XAI interaction (RQ5). No significant interactions.}
\label{tab:rq5_int_xai}
\end{table}

\begin{table}[H]
\centering
\small
\resizebox{\columnwidth}{!}{%
\begin{tabular}{lcccc}
\toprule
\textbf{Dimension} & \textbf{F} & \textbf{p} & \textbf{$\omega^2$} & \textbf{Sig.} \\
\midrule
Accuracy & 0.66 & .876 & .000 & -- \\
Lay Relevancy & 0.75 & .787 & .000 & -- \\
Expert Relevancy & 0.42 & .990 & .000 & -- \\
Helpfulness & 0.57 & .940 & .000 & -- \\
\bottomrule
\end{tabular}%
}
\caption{Strategy $\times$ Model interaction (RQ5). No significant interactions.}
\label{tab:rq5_int_model}
\end{table}

\paragraph{Ranking Evaluation.} To address potential positional bias in absolute scoring \citep{gu2024llmasjudge}, we conducted a supplementary \emph{listwise} ranking evaluation. Both judges ranked all 8 strategies for 60 matched sets ($N=120$ rankings total). Table~\ref{tab:rq5_ranking} shows average ranks and frequency of best/worst rankings.

\begin{table}[H]
\centering
\small
\resizebox{\columnwidth}{!}{%
\begin{tabular}{lccccc}
\toprule
\textbf{Strategy} & \textbf{Avg Rank} & \textbf{SD} & \textbf{\#Best} & \textbf{\#Worst} & \textbf{G-Eval} \\
\midrule
Self-consistency & 2.86 & 1.81 & 37 & 2 & 4.44 \\
Reflexion & 3.41 & 1.74 & 19 & 2 & 4.35 \\
CoT Zero-shot & 3.71 & 1.78 & 16 & 1 & 4.17 \\
Zero-shot & 3.92 & 2.02 & 16 & 6 & 4.35 \\
Role-based & 4.14 & 2.03 & 14 & 6 & 4.32 \\
Few-shot & 5.86 & 1.98 & 6 & 22 & 4.25 \\
CoT Few-shot & 6.02 & 2.01 & 5 & 35 & 4.01 \\
Meta-prompting & 6.07 & 2.05 & 6 & 45 & 4.01 \\
\bottomrule
\end{tabular}%
}
\caption{Ranking evaluation results (RQ5). Lower rank = better. $N=119$ rankings (combined from 2 judges).}
\label{tab:rq5_ranking}
\end{table}

\paragraph{Friedman Test.} Non-parametric repeated-measures test for ranking data: $\chi^2(7)=230.3$, $p<.001$. Strategy rankings are significantly different.

\paragraph{G-Eval vs Ranking Correlation.} Spearman correlation between G-Eval average scores (Table~\ref{tab:strategy_results}) and average ranks (Table~\ref{tab:rq5_ranking}): $\rho=-.857$, $p=.007$. The strong negative correlation (higher G-Eval score $\leftrightarrow$ lower/better rank) validates our pointwise G-Eval evaluation methodology.

\paragraph{Inter-Judge Agreement.} Kendall's $W=0.72$ for ranking agreement across 60 sets, indicating substantial agreement between judges.

\end{document}